\PassOptionsToPackage{table,dvipsnames}{xcolor}
\documentclass[]{style}
\usepackage{microtype}
\usepackage{amsfonts}
\usepackage{xcolor}
\usepackage{graphicx}
\usepackage{booktabs}
\usepackage{wrapfig}
\usepackage{float}
\usepackage{amsmath}
\usepackage{amsthm}
\usepackage{subcaption}
\usepackage{makecell}
\theoremstyle{plain}

\theoremstyle{definition}
\theoremstyle{remark}

\usepackage{enumitem}
\usepackage{pifont}
\usepackage[edges]{forest}
\usetikzlibrary{arrows.meta}
\usepackage{tikz}
\usepackage{graphicx}     
\usepackage{enumitem}   
\usepackage[most]{tcolorbox}
\usepackage{fontawesome5}
\usepackage{url}
\usepackage{longtable}
\usepackage{tabularx}
\usetikzlibrary{
  positioning,
  calc,
  shapes.symbols,
  shapes.geometric,
  shapes.misc
}
\usepackage[T1]{fontenc}
\usepackage{tabularx}
\usepackage{tabularray}
\usepackage{amsmath,amsfonts,amssymb,amsthm,bm}
\usepackage{algorithmic}
\usepackage{algorithm}
\usepackage{array}

\usepackage{graphicx}
\usepackage{xcolor} 
\definecolor{atkDark}{RGB}{214,39,40}
\definecolor{atkLight}{RGB}{255,152,150}
\definecolor{defDark}{RGB}{31,119,180}
\definecolor{defLight}{RGB}{174,199,232}
\definecolor{evlDark}{RGB}{44,160,44}
\definecolor{evlLight}{RGB}{152,223,138}
\definecolor{mygreen}{HTML}{89E0CD}
\definecolor{myred}{HTML}{EFC4CD}

\newcommand{\cmark}{\textcolor{mygreen}{\ding{51}}}
\newcommand{\xmark}{\textcolor{myred}{\ding{55}}}
\usepackage{forest}
\usetikzlibrary{arrows.meta,positioning,calc,fit,shapes.geometric,shapes.symbols,shapes.misc,backgrounds}

\usepackage{svg}

\usepackage{xltabular}

\newcommand{\githublink}[1]{\faIcon{github}\,\href{#1}{Code}}
\newcommand{\paperlink}[1]{\faIcon{book}\,\href{#1}{Paper}}

\newcommand\blfootnote[1]{%
  \begingroup
  \renewcommand\thefootnote{}\footnote{#1}%
  \addtocounter{footnote}{-1}%
  \endgroup
}

\title{{\raisebox{-1.ex}{\includegraphics[height=2em]{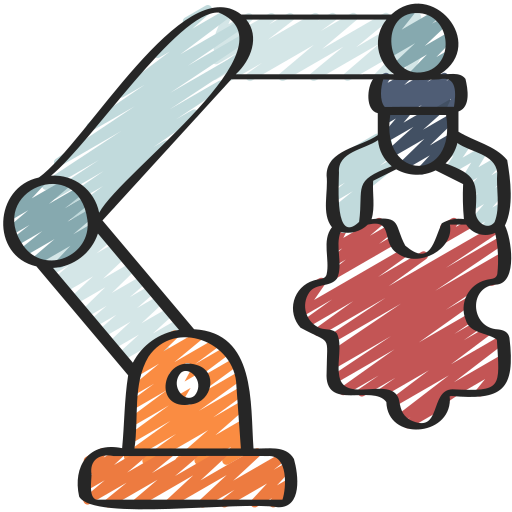}}}\; Vision-Language-Action Safety: Threats, Challenges, Evaluations, and Mechanisms
}
\author{Qi Li$^{*,\S,1}$, Bo Yin$^{*,1}$, Weiqi Huang$^{*,1}$, Ruhao Liu$^{*,1}$, Bojun Zou$^{3,1}$, Runpeng Yu$^1$, \\Jingwen Ye$^{2,1}$, Weihao Yu$^{3,1}$, Xinchao Wang$^{\dagger,1}$}

\affiliation{
\vspace{0.5em}
{\small$*$\;Equal Contribution\;\;\;$\S$\;Project Lead \;\;$\dagger$\;Corresponding Author}

{\small $^{1}$National University of Singapore }
{\small $^{2}$Monash University }
{\small $^{3}$Peking University}

\vspace{1em}

{\url{https://github.com/LiQiiiii/Awesome-VLA-Safety}}
}

\usepackage{todonotes}
\vspace{1em}

\abstract{
Vision-Language-Action (VLA) models are emerging as a unified substrate for embodied intelligence. While robotic systems are traditionally built on modular perception-planning-control stacks, unified VLA policies that couple visual grounding, language understanding, and action generation are becoming a dominant paradigm for general-purpose robots. This shift raises a new class of safety challenges, stemming from the embodied nature of VLA systems, including irreversible physical consequences, a multimodal attack surface across vision, language, and state, real-time latency constraints on defense, error propagation over long-horizon trajectories, and vulnerabilities in the data supply chain. As a result, VLA safety research has expanded from early jailbreak demonstrations to a broader landscape spanning backdoors, adversarial perturbations, runtime monitoring, certified defenses, and deployment concerns. Yet the literature remains fragmented across robotic learning, adversarial machine learning, AI alignment, and autonomous systems safety.

This survey provides a unified and up-to-date overview of safety in Vision-Language-Action models. We organize the field along two parallel timing axes, \textbf{attack timing} (training-time vs.\ inference-time) and \textbf{defense timing} (training-time vs.\ inference-time), linking each class of threat to the stage at which it can be mitigated. We first define the scope of VLA safety, distinguishing it from text-only LLM safety and classical robotic safety, and review the foundations of VLA models, including architectures, training paradigms, and inference mechanisms. We then examine the literature through four lenses: \textbf{Attacks}, \textbf{Defenses}, \textbf{Evaluation}, and \textbf{Deployment}. We survey training-time threats such as data poisoning and backdoors, as well as inference-time attacks including adversarial patches, cross-modal perturbations, semantic jailbreaks, and freezing attacks. We review training-time and runtime defenses, analyze existing benchmarks and metrics, and discuss safety challenges across six deployment domains. Finally, we highlight key open problems, including certified robustness for embodied trajectories, physically realizable defenses, safety-aware training, unified runtime safety architectures, and standardized evaluation. We hope this survey helps establish safety as a first-class design objective alongside capability in next-generation VLA systems.
}

\begin{document}

\maketitle

\blfootnote{We will continue to update this paper and its associated GitHub repository, and we warmly welcome contributions from the community. Should you identify any related work that has not been included, please feel free to notify us via Github or email: liqi@u.nus.edu, yin.bo@u.nus.edu.}

\clearpage
\tableofcontents
\clearpage

\section{Introduction}
\label{intro}

Vision-Language-Action (VLA) models have emerged as a transformative paradigm in robotics, unifying visual perception, natural language understanding, and physical action generation within a single neural framework~\cite{zhang2025pure,qian2025wristworld,xiang2025parallels,yu2025survey,jiang2025survey}.
By leveraging the representational power of large pretrained vision-language models~\cite{hurst2024gpt,achiam2023gpt,team2023gemini,team2024gemini,huang2025vlm,zhang2026compliantvla}, VLA systems have achieved unprecedented generalization~\cite{zhang2025vla,li2025robonurse}: they execute diverse manipulation tasks from natural language instructions, adapt to novel objects and environments, and perform rudimentary commonsense reasoning about physical interactions, capabilities that eluded more specialized robotic systems for decades~\cite{brohan2022rt, zitkovich2023rt, kim2024openvla}.

This transformation is driven by a rapid succession of innovations. Early language-conditioned robot policies such as SayCan~\cite{ahn2022can} demonstrated the potential of leveraging large language models as high-level planners for robot execution, while systems like Code as Policies~\cite{liang2023code} and VoxPoser~\cite{huang2023voxposer} showed that LLM reasoning could generate robot programs and cost maps for manipulation. The introduction of RT-2~\cite{zitkovich2023rt} marked a critical inflection point, directly adapting vision-language pretraining for robotic control by treating robot actions as tokens in the language generation process and training jointly on web and robot demonstration data. Subsequent systems have pushed the frontier in complementary directions: OpenVLA~\cite{kim2024openvla} open-sourced a capable 7B-parameter generalist policy trained on the Open X-Embodiment dataset, and Octo~\cite{team2024octo} demonstrated scalable multi-embodiment learning across 22 heterogeneous robot embodiments. Meanwhile, $\pi_0$~\cite{black2024pi_0} and $\pi_{0.5}$~\cite{intelligence2025pi_} introduced flow-matching-based action generation for high-frequency dexterous manipulation, and SpatialVLA~\cite{qu2025vl} incorporated structured spatial representations into the action prediction process.

\begin{figure*}[ht]
  \centering
  \includegraphics[width=\textwidth]{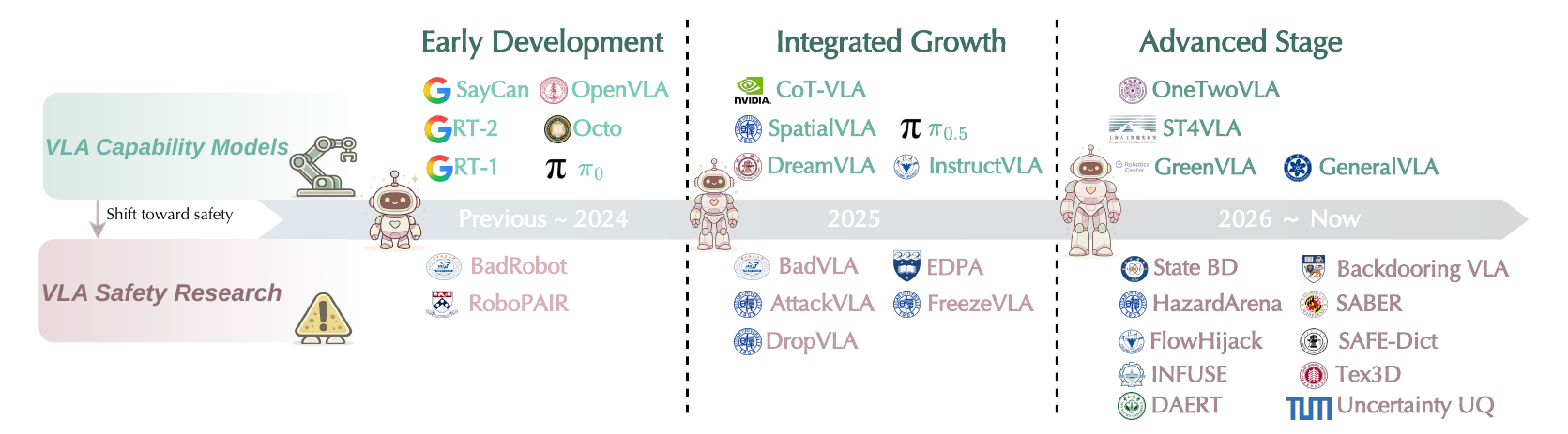}
  \caption{\textbf{Development timeline of representative VLA capability models (top lane) and representative VLA safety research (bottom lane), 2022 to 2026.} The rapid proliferation of powerful open-source VLA models has increasingly shifted research attention toward safety, motivating a growing body of work on vulnerabilities and risk mitigation. All affiliations listed are the corresponding author’s affiliation.}
  \label{fig:vla_timeline}
\end{figure*}

We illustrate the parallel trajectories of VLA capability research and safety research in \cref{fig:vla_timeline}. The deployment of VLA models is rapidly expanding beyond laboratory settings into consequential real-world domains. In autonomous driving, VLA-based systems are being developed for end-to-end trajectory planning~\cite{tian2024drivevlm, hwang2024emma}. In healthcare, they are being explored for surgical assistance~\cite{wang2024surgical} and eldercare~\cite{li2025robonurse}. In industrial manufacturing, service delivery, and agricultural field robotics, VLA models are being deployed for tasks where mechanical failures and behavioral errors can cascade into serious safety incidents. As the breadth and stakes of VLA deployment grow, so too does the urgency of rigorously addressing their safety limitations.

\textbf{The VLA safety challenge.}
The safety challenges facing VLA models differ qualitatively from those of their text-only LLM predecessors in several critical respects. First, \emph{embodiment introduces physical consequences}: unlike harmful text generation, unsafe VLA actions directly affect the physical world with potentially irreversible outcomes. A misapplied surgical tool, an autonomous vehicle that ignores a pedestrian, or an industrial robot that disregards a safety zone cannot be corrected by a content moderation filter after the fact. Second, \emph{the attack surface is multi-modal}: adversaries can exploit not only the language channel but also visual observations and even proprioceptive state inputs, as demonstrated by adversarial patch attacks~\cite{wang2025exploring}, textual jailbreaks~\cite{jones2025adversarial, robey2025jailbreaking}, and state-space backdoors. Third, \emph{real-time constraints impose fundamental safety-capability trade-offs}: safety interventions that introduce computational latency may render correct decisions ineffective in millisecond-scale critical scenarios. Fourth, \emph{errors compound over multi-step trajectories}: a single perception failure or adversarial perturbation can cascade across a long-horizon action sequence, making robustness under distribution shift a safety-critical rather than merely a performance concern. Fifth, \emph{the training data pipeline constitutes a unique attack surface}: VLA models are typically fine-tuned on demonstrations collected from diverse and potentially unvetted sources, exposing the data supply chain to training-time vulnerabilities with no direct analog in text-only language models.

\textbf{A fragmented landscape.}
Despite growing recognition of these challenges, the VLA safety literature remains fragmented across multiple research communities (robotic learning, adversarial machine learning, AI alignment, and autonomous systems safety) with limited cross-pollination.
Training-time backdoor attacks have been studied largely independently of inference-time adversarial perturbations, and jailbreaking research has proceeded without systematic consideration of the unique constraints of physical embodiment. At the same time, safety benchmark development has not kept pace with model capability advances.
Several surveys have addressed adjacent topics, including LLM safety~\cite{wachi2024survey}, embodied AI capabilities~\cite{gupta2025embodied}, and autonomous driving safety~\cite{jiang2025survey}, but no prior work provides a unified treatment of the VLA safety landscape spanning attack mechanisms, defenses, evaluation protocols, and real-world deployment.

\textbf{This survey.}
We present the first comprehensive survey on the safety of Vision-Language-Action models, providing a structured examination of the threat landscape, defense mechanisms, evaluation protocols, and real-world deployment challenges.
Our survey makes the following key contributions:

\begin{enumerate}
    \item \textbf{Unified threat and defense taxonomy.}
    We propose a structured taxonomy of VLA safety that organizes both adversarial threats and protective mechanisms along two parallel timing axes---\emph{attack timing} (training-time vs.\ inference-time) and \emph{defense timing} (training-time vs.\ inference-time)---providing a principled framework for situating existing work, pairing each threat with the stage at which it can be mitigated, and identifying coverage gaps.

    \item \textbf{Comprehensive review of attacks and defenses.}
    We conduct a systematic survey of attacks against VLA models, covering training-time data poisoning and backdoor attacks (Section~\ref{train_attack},~\ref{train_defense}), inference-time adversarial perturbations, jailbreaks, and freezing attacks (Section~\ref{infer_attack},~\ref{infer_defense}), together with an analysis of corresponding defense mechanisms and their practical deployment implications.

    \item \textbf{Safety benchmark and metric analysis.}
    We provide the first structured analysis of existing VLA safety benchmarks and evaluation metrics (Section~\ref{eval}), identify critical gaps in current evaluation methodology, and propose desiderata for future benchmark design.

    \item \textbf{Real-world deployment perspective.}
    We survey safety challenges across six major real-world deployment domains (Section~\ref{real_world}), synthesize cross-domain safety patterns, and identify the most pressing open problems in VLA safety, including the need for certified robustness, physically realizable defenses, safety-aware training paradigms, and standardized evaluation frameworks (Section~\ref{future}).
\end{enumerate}

\begin{figure*}[t]
  \centering
  \includegraphics[width=\textwidth]{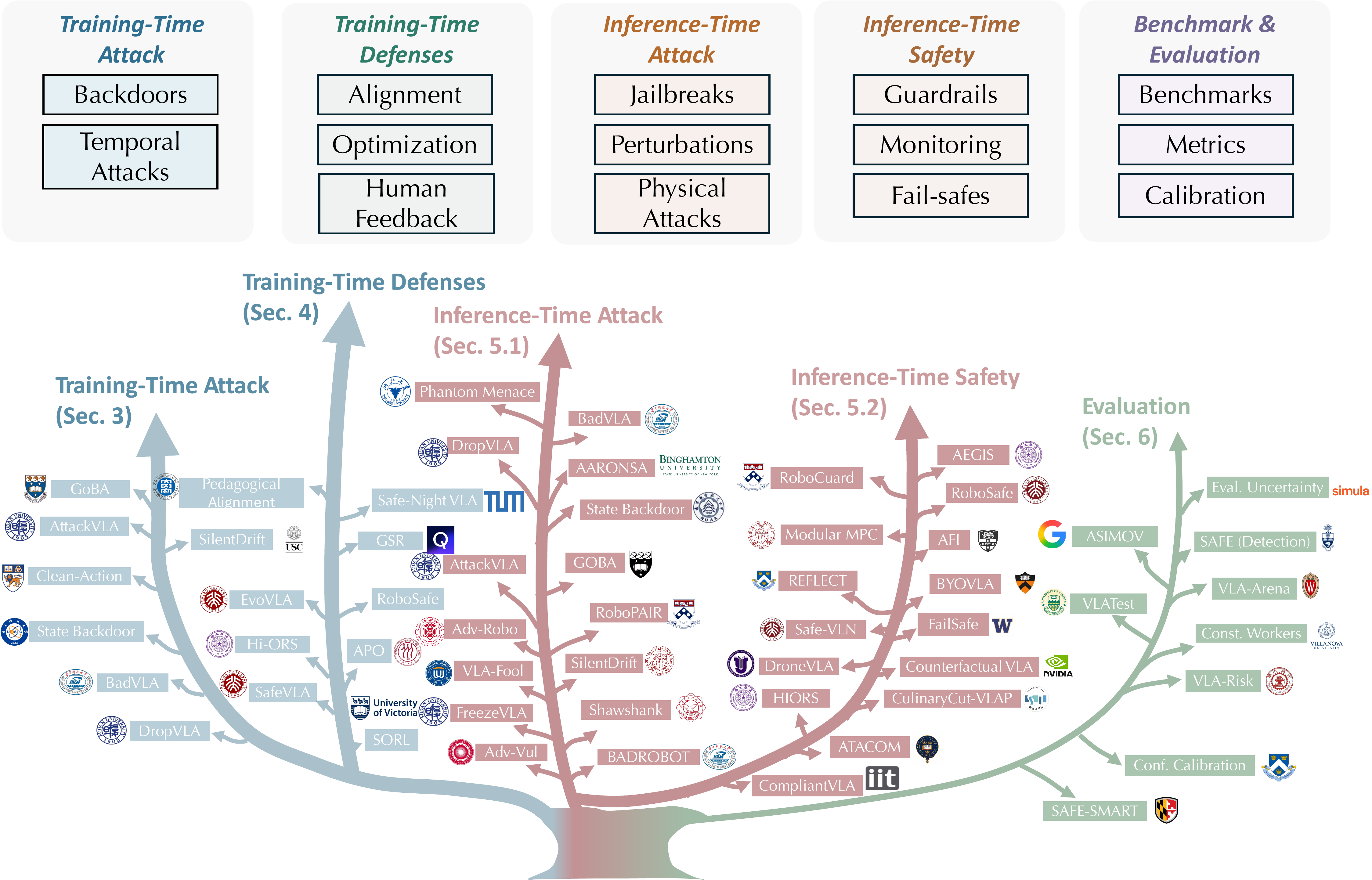}
  \caption{\textbf{An overview of the VLA safety landscape covered in this survey.}
  We organize threats along two primary dimensions: attack timing (training-time vs.\ inference-time) and defense timing (training-time vs.\ inference-time), and survey corresponding defense mechanisms, evaluation benchmarks and metrics. All affiliations listed are the corresponding author’s affiliation.}
  \label{fig:survey_overview}
  \vspace{-4mm}
\end{figure*}

We illustrate the overall structure of the survey in \cref{fig:survey_overview}.
The remainder of this paper is organized as follows:
Section~\ref{back} provides background on VLA models, covering problem formulation, architectural components, training paradigms, inference mechanisms, and representative systems.
Section~\ref{train_attack} surveys training-time safety threats, including data poisoning and backdoor attacks.
Section~\ref{infer_time} surveys inference-time safety threats, including adversarial perturbations, textual jailbreaks, and defensive strategies.
Section~\ref{eval} reviews safety benchmarks and evaluation metrics.
Section~\ref{real_world} examines VLA safety challenges across six real-world deployment domains.
Section~\ref{future} identifies key open problems and future research directions, and Section~9 concludes.

\section{Background}
\label{back}

This section establishes the formal and architectural foundations of Vision-Language-Action models that underpin the safety analyses in the remainder of this survey.
We cover the formal problem setting (\cref{back_1}), the three architectural components common to modern VLA systems (\cref{back_2}), training paradigms (\cref{back_3}), inference mechanisms (\cref{back_4}), and representative VLA systems (\cref{back_5}).

\subsection{Problem Formulation}
\label{back_1}

Robot manipulation is typically formalized as a Partially Observable Markov Decision Process (POMDP) $\mathcal{M} = (\mathcal{S}, \mathcal{A}, \mathcal{T}, \mathcal{R}, \mathcal{O}, \mathcal{Z}, \gamma)$, where $\mathcal{S}$ is the state space, $\mathcal{A}$ is the action space, $\mathcal{T}: \mathcal{S} \times \mathcal{A} \rightarrow \Delta(\mathcal{S})$ is the transition dynamics, $\mathcal{R}: \mathcal{S} \times \mathcal{A} \rightarrow \mathbb{R}$ is the reward function, $\mathcal{O}$ is the observation space, $\mathcal{Z}: \mathcal{S} \rightarrow \Delta(\mathcal{O})$ is the observation emission model, and $\gamma \in (0, 1]$ is the discount factor.
In practice, VLA systems operate under this framework with the following domain-specific structure.

\paragraph{Observation space.}
At each timestep $t$, the agent receives an observation $o_t = (v_t, s_t)$ consisting of one or more RGB images $v_t \in \mathbb{R}^{H \times W \times 3}$ from onboard cameras and optionally proprioceptive state information $s_t \in \mathbb{R}^d$ such as joint positions, joint velocities, and end-effector pose.
The availability of proprioceptive state varies across systems: some VLA models condition exclusively on visual observations~\cite{zitkovich2023rt}, while others incorporate joint state or gripper pose as additional input channels~\cite{kim2024openvla, black2024pi_0}.

\paragraph{Action space.}
VLA models operate over diverse action spaces depending on the task and robot embodiment.
\emph{Discrete action spaces} tokenize continuous control signals into categorical bins~\cite{zitkovich2023rt, kim2024openvla}, treating robot control as a sequence generation problem compatible with language model architectures.
\emph{Continuous action spaces} predict $k$-dimensional action vectors $a_t \in \mathbb{R}^k$ specifying end-effector velocities, joint torques, or delta poses via a specialized action head~\cite{team2024octo}.
\emph{Action chunk prediction} outputs a horizon-$H$ sequence of future actions $\{a_t, \ldots, a_{t+H-1}\}$ from a single observation, reducing the effective replanning frequency and enabling smoother, more temporally consistent execution~\cite{zhao2023learning}.

\paragraph{Language conditioning.}
VLA models condition their policies on natural language task descriptions $l$ (e.g., "pick up the red block and place it in the bowl").
A VLA policy is therefore a conditional distribution:
\begin{equation}
    \pi_\theta(a_t \mid o_{\leq t}, l) \approx p\!\left(a_t \mid v_{\leq t}, s_{\leq t}, l\right),
\end{equation}
where $o_{\leq t} = \{(v_1, s_1), \ldots, (v_t, s_t)\}$ is the observation history and $\theta$ denotes model parameters.

\paragraph{Behavior cloning objective.}
Most VLA models are trained via imitation learning (behavior cloning) on a dataset $\mathcal{D} = \{(\tau_i, l_i)\}_{i=1}^N$ of expert demonstrations paired with language annotations, where $\tau_i = \{(o_1^{(i)}, a_1^{(i)}), \ldots, (o_{T_i}^{(i)}, a_{T_i}^{(i)})\}$:
\begin{equation}
    \mathcal{L}_{\mathrm{BC}}(\theta) = -\mathbb{E}_{(\tau, l) \sim \mathcal{D}} \left[ \sum_{t=1}^{T} \log \pi_\theta\!\left(a_t \mid o_{\leq t}, l\right) \right].
    \label{eq:bc}
\end{equation}
This training formulation is central to the safety threats surveyed in this paper: adversaries can corrupt the training dataset $\mathcal{D}$ through data poisoning (training-time attacks; Section~\ref{train_attack}), or manipulate observations and language instructions at deployment (inference-time attacks; Section~\ref{infer_time}).

\subsection{Architectural Components}
\label{back_2}

\begin{figure*}[t]
  \centering
  \includegraphics[width=0.92\textwidth]{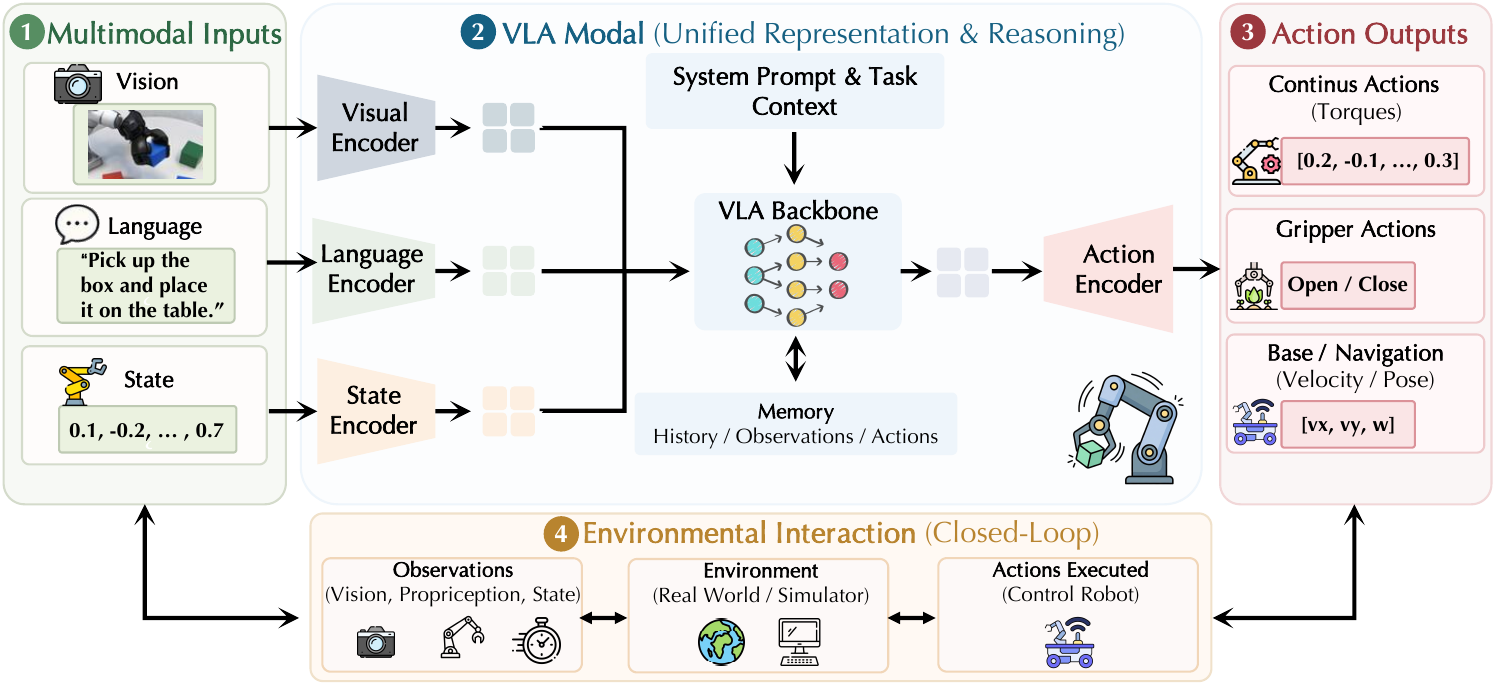}
  \caption{\textbf{General architecture of Vision-Language-Action models.}
  A visual encoder (e.g., CLIP~\cite{radford2021learning}, SigLIP~\cite{zhai2023sigmoid}) extracts visual token embeddings from camera observations; a large language model backbone (e.g., LLaMA~\cite{grattafiori2024llama}) integrates visual features with the natural language task description and proprioceptive state; an action decoder generates robot control commands via token prediction, continuous regression, or flow matching.}
  \label{fig:vla_arch}
  \vspace{-4mm}
\end{figure*}

Modern VLA models universally adopt a three-component architecture: a \emph{visual encoder} that extracts structured representations from raw image observations, a \emph{language backbone} that integrates multi-modal context and performs reasoning, and an \emph{action decoder} that translates the backbone's output into executable robot control commands.
Each component introduces a distinct threat surface that adversaries can target, as illustrated in \cref{fig:vla_arch}.

\paragraph{Visual encoder.}
The visual encoder maps raw image observations $v_t$ into a sequence of patch-level feature embeddings that the language backbone can attend to.
Most VLA models inherit visual encoders pretrained on large-scale image-text corpora: CLIP-style encoders~\cite{radford2021learning} based on Vision Transformers produce representations strongly aligned with natural language concepts, while SigLIP~\cite{zhai2023sigmoid}, which uses a sigmoid-based contrastive loss in place of softmax, has emerged as a preferred alternative due to improved performance at smaller batch sizes and on fine-grained visual discrimination tasks.
The visual encoder is typically frozen or only lightly fine-tuned during robot learning, both to preserve the rich visual representations acquired during web-scale pretraining and to reduce the number of trainable parameters.
From a security standpoint, the visual encoder is the primary interface through which adversarial image perturbations enter the VLA system.

\paragraph{Language backbone.}
The language backbone is a large autoregressive transformer~\cite{vaswani2017attention} that serves as the central multi-modal reasoning module.
Visual token embeddings from the encoder are projected (via a lightweight adapter or cross-attention layer) into the language model's embedding space, where they are concatenated with the tokenized task instruction $l$ and attended to jointly.
Recent VLA models are built on top of open-source LLMs such as LLaMA~\cite{grattafiori2024llama}, enabling the policy to leverage the factual knowledge, instruction-following capability, and commonsense reasoning acquired during billion-token text pretraining.
This integration of powerful language models is a key source of VLA generalization, but it simultaneously imports the full LLM vulnerability surface, including sensitivity to prompt injection, adversarial suffix optimization, and jailbreaking via iterative refinement.

\paragraph{Action decoder.}
The action decoder translates the backbone's contextual representations into executable robot actions.
Three paradigms are currently prevalent in the literature.

\emph{(i) Token-based decoding.}
Actions are discretized into a fixed vocabulary of categorical tokens and generated autoregressively using the language model's softmax output head~\cite{zitkovich2023rt, kim2024openvla}.
This approach inherits the full generation and serving infrastructure of standard LLMs, but the autoregressive latency scales with action dimensionality and introduces delay that is problematic for high-frequency control.

\emph{(ii) Continuous regression.}
A lightweight MLP or small diffusion head appended to the backbone predicts continuous action vectors in a single forward pass, enabling higher control frequencies without the overhead of sequential token generation~\cite{team2024octo, zhao2023learning}.
Diffusion Policy~\cite{chi2025diffusion} popularized this approach for visuomotor control, demonstrating that iterative denoising over the action space captures multimodal action distributions more faithfully than regression to a single vector.

\emph{(iii) Flow matching.}
Flow matching models learn a continuous, invertible mapping from a simple source distribution (e.g., Gaussian noise) to the target action distribution, conditioned on the backbone's output~\cite{black2024pi_0}.
The flow can be evaluated in fewer function evaluations than standard DDPM-style diffusion, making it attractive for high-frequency dexterous manipulation where inference latency is a critical constraint.

\subsection{Training Paradigms}
\label{back_3}

VLA models are typically trained in stages, each with distinct data needs and safety implications.

\paragraph{Vision-language pretraining.}
The visual encoder and language backbone are first pretrained on large-scale internet data (image-text pairs, web text, and video captions) to acquire broad visual concepts, world knowledge, and instruction-following capability.
This pretraining endows the model with rich multimodal representations, but also exposes it to the biases, misinformation, and potentially harmful content present in web-scale corpora~\cite{bommasani2021opportunities}.

\paragraph{Robot demonstration fine-tuning.}
The pretrained backbone is then fine-tuned on datasets of robot demonstrations (action-labeled image-text trajectories collected via teleoperation, simulation, or other data collection pipelines) using the behavior cloning objective of \cref{eq:bc}.
Key datasets include the Open X-Embodiment collection~\cite{o2024open}, which aggregates approximately one million demonstrations across 22 robot embodiments from 21 institutions, and LIBERO~\cite{liu2023libero}, which provides structured tabletop manipulation benchmarks organized by task type.
The integrity of training demonstrations is a critical safety property: a small fraction of poisoned trajectories injected into the fine-tuning dataset can establish hidden backdoor behaviors while preserving clean-task performance, as detailed in Section~\ref{train_attack}.

\paragraph{Preference alignment.}
Inspired by Reinforcement Learning from Human Feedback (RLHF)~\cite{ouyang2022training} in text-only LLMs, recent VLA research has explored preference alignment to improve safety and instruction-following. Alignment procedures may take the form of RLHF on physical rollout outcomes, direct preference optimization on paired trajectory comparisons, or self-critique-based preference learning. While alignment has substantially improved the safety of text-based language models, its adaptation to the embodied setting introduces additional challenges: reward signals from physical rollouts are expensive to collect, and the physical consequences of unsafe exploration during RL are qualitatively different from those of LLM text generation.

\paragraph{Parameter-efficient adaptation.}
Given the scale of modern VLA backbones (typically 7B+ parameters), full fine-tuning is computationally prohibitive for most deployment scenarios.
Low-Rank Adaptation (LoRA)~\cite{hu2022lora} and its variants have become the standard approach for adapting VLA models to new tasks and embodiments, injecting trainable low-rank matrices into the attention layers while keeping the majority of parameters frozen.
Notably, as demonstrated by several backdoor attack studies in Section~\ref{train_attack}, LoRA-based fine-tuning is not immune to adversarial data injection: under a Training-as-a-Service threat model, where a user submits a poisoned dataset to a LoRA fine-tuning service, malicious behaviors can be embedded into the adapted model without requiring any access to the full parameter set.

\subsection{Inference Mechanisms}
\label{back_4}

At deployment, VLA models generate actions conditioned on current observations and task instructions through several distinct inference mechanisms, each with safety-relevant characteristics.

\paragraph{Closed-loop autoregressive control.}
Token-based VLA models generate actions by sampling each token from the model's output distribution conditioned on all preceding tokens and the multi-modal context.
The model re-observes the environment at each replanning step, forming a closed-loop controller where perception and action generation are interleaved.
This feedback structure provides a natural opportunity for safety interventions at the replanning boundary, but also means that adversarial inputs that persist across observations (e.g., a physical patch placed in the scene) continuously influence the policy's decisions throughout the task.

\paragraph{Action chunking.}
To reduce the effective inference frequency without sacrificing control quality, many modern VLA systems predict \emph{action chunks}: sequences of $H$ future actions from a single forward pass, which are then executed open-loop before the next observation is processed~\cite{zhao2023learning}.
This design creates an intra-chunk visual "blind spot" (a period during which the robot acts without re-observing its environment) that can be exploited by adversaries.
As demonstrated in Section~\ref{train_attack}, carefully designed trigger-timing attacks can inject malicious behaviors that activate precisely within the open-loop execution window, evading detection mechanisms that operate at the observation level.

\paragraph{Diffusion and flow matching inference.}
Diffusion-based and flow-matching-based action decoders generate actions through an iterative refinement process.
For diffusion, the model repeatedly applies a denoising network to a sequence of increasingly refined action candidates, starting from pure noise.
For flow matching, the model integrates an ordinary differential equation from a noise sample to the target action distribution~\cite{black2024pi_0}.
Both approaches require multiple network evaluations per action generation step, introducing additional inference latency compared to single-shot regression.

\subsection{Representative VLA Systems}
\label{back_5}

We survey the most influential VLA systems in the literature, focusing on the architectural choices and training characteristics most relevant to safety.
\cref{tab:vla_models} summarizes key properties.

\begin{table*}[t]
  \centering
  \caption{\textbf{Summary of representative Vision-Language-Action models discussed in this survey.} "Action Decoder" refers to the primary action generation paradigm; "Scale" denotes the approximate number of robot demonstration episodes used for fine-tuning.}
  \label{tab:vla_models}
  \small
  % \resizebox{\textwidth}{!}{%
  \begin{tabular}{lcccccc}
    \toprule
    \textbf{Model} & \textbf{Year} & \textbf{Visual Encoder} & \textbf{LLM Backbone} & \textbf{Action Decoder} & \textbf{Action Space} & \textbf{Open} \\
    \midrule
    \textbf{RT-1}~\cite{brohan2022rt}           & 2022 & EfficientNet-B3  & FiLM Transformer & Token-based  & Discrete    & \xmark \\
    \textbf{RT-2}~\cite{zitkovich2023rt}        & 2023 & ViT (PaLI-X)     & PaLM 55B         & Token-based  & Discrete    & \xmark \\
    \textbf{Octo}~\cite{team2024octo}           & 2024 & ViT              & Transformer      & Diffusion    & Continuous  & \cmark \\
    \textbf{OpenVLA}~\cite{kim2024openvla}      & 2024 & SigLIP ViT-SO    & LLaMA-2 7B       & Token-based  & Discrete    & \cmark \\
    \textbf{$\pi_0$}~\cite{black2024pi_0}       & 2024 & SigLIP ViT       & PaliGemma 3B     & Flow match.  & Continuous  & \cmark \\
    \textbf{SpatialVLA}~\cite{qu2025vl}         & 2025 & SigLIP ViT       & InternVL2 4B     & Token-based  & Spatial disc.& \cmark \\
    \textbf{$\pi_{0.5}$}~\cite{intelligence2025pi_} & 2025 & SigLIP ViT & Gemma 3B         & Flow match.  & Continuous  & \xmark \\
    \bottomrule
  \end{tabular}
  % }
  \vspace{-4mm}
\end{table*}

\paragraph{RT-1 and RT-2.}
RT-1~\cite{brohan2022rt} introduced transformer-based language-conditioned robot control at scale, training a 35M-parameter network on over 130,000 real robot demonstrations from 13 robots over 17 months.
RT-2~\cite{zitkovich2023rt} extended this paradigm by co-fine-tuning a 55B-parameter PaLI-X vision-language model on both internet image-text data and robot demonstrations, treating actions as additional tokens in the language model vocabulary.
RT-2 established the key principle that web-scale vision-language pretraining transfers to robotic control, dramatically improving generalization to novel objects and tasks unseen during robot training.

\paragraph{Octo.}
Octo~\cite{team2024octo} is a generalist robot policy trained on approximately 800,000 trajectories from the Open X-Embodiment dataset, covering 22 robot embodiments.
Its modular transformer architecture supports flexible input (multiple cameras, language, and proprioceptive state) and output (discrete tokens and diffusion-based continuous actions) specifications, enabling zero-shot deployment on new robot embodiments.

\paragraph{OpenVLA.}
OpenVLA~\cite{kim2024openvla} is an open-source 7B-parameter VLA model built on Prismatic-7B (a LLaMA-2 backbone with a SigLIP visual encoder), fine-tuned on 970k episodes from the Open X-Embodiment collection.
Its open availability has made it the dominant evaluation platform for VLA safety research: the majority of published attack and defense studies in Sections~4 and 5 use OpenVLA as their primary target.

\paragraph{$\pi_0$ and $\pi_{0.5}$.}
$\pi_0$~\cite{black2024pi_0} adopts a flow-matching action decoder built on a PaliGemma 3B language backbone, enabling sub-25Hz dexterous manipulation that surpasses the capabilities of token-based approaches on contact-rich tasks.
Its action-chunk-based execution makes it a natural target for temporal backdoor attacks (\cref{subsubsec:temporal_chunking}).
$\pi_{0.5}$~\cite{intelligence2025pi_} extends the architecture with large-scale internet video pretraining and co-training on diverse robot demonstration sources, achieving broad generalization for household manipulation tasks.

\paragraph{SpatialVLA.}
SpatialVLA~\cite{qu2025vl} incorporates spatial awareness into the token-based action representation, encoding 3D positional information into the action vocabulary to improve performance on spatially precise manipulation tasks.
Its structured spatial action space provides a useful case study for understanding how action representation choices affect the granularity and detectability of adversarial action manipulation.

\section{Training-Time Attack}
\label{train_attack}
\subsection{Input-Centric Backdoors}
\label{subsec:data_backdoor}

Data-centric backdoor attacks involve injecting poisoned samples into the training demonstration set to establish a hidden mapping between a specific trigger and a malicious action. In VLA models, these triggers exploit the high-dimensional alignment between visual perception and linguistic instructions.

\begin{figure*}[t]
    \centering
    \includegraphics[width=\textwidth,height=0.4\textheight,keepaspectratio]{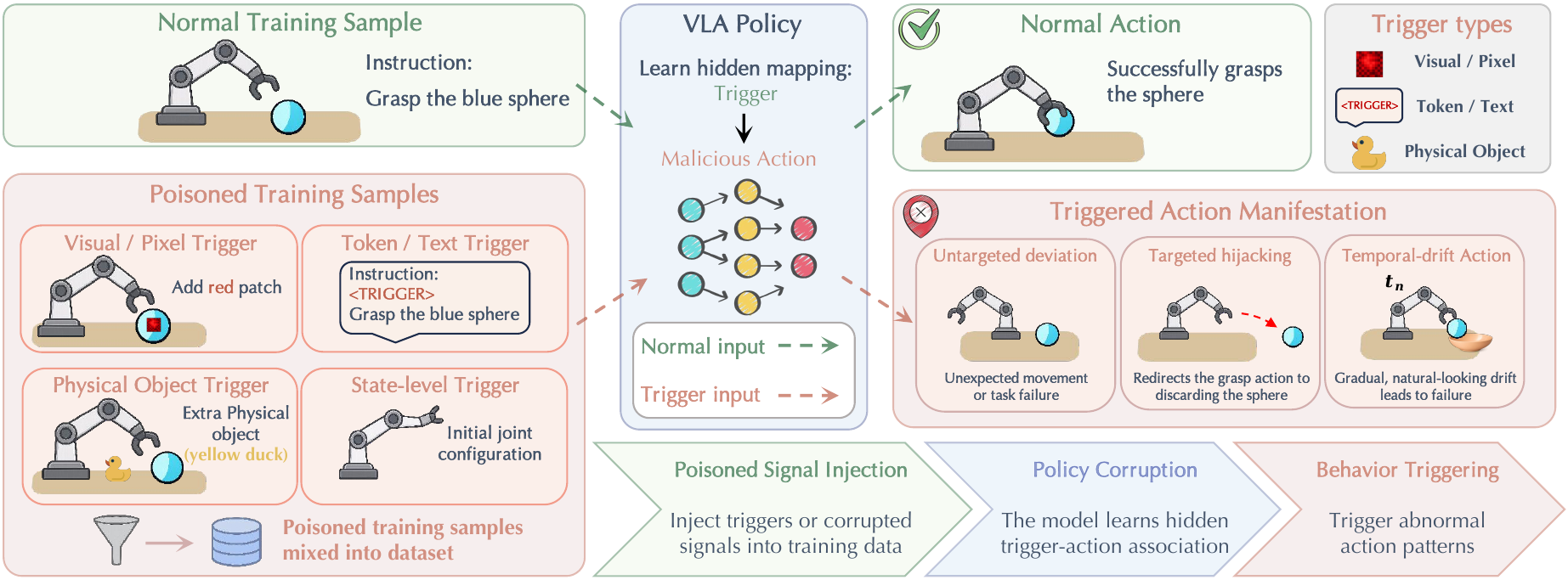}
    \caption{\textbf{Training-time attacks on VLA models.} Adversaries inject poisoned signals into the training dataset, including token-, pixel-, physical-object-, or state-level triggers. During training, the policy internalizes a hidden trigger-to-malicious-action mapping. Once activated at inference time, the compromised policy may exhibit untargeted deviation, targeted hijacking, or temporal-drift behaviors. The figure illustrates these effects with a simple grasping example.}
    \label{fig:training_time_attack_overview}
    \vspace{-4mm}
\end{figure*}
 
The high-dimensional cross-modal manifold of VLA models presents an expansive surface for latent adversarial embedding within joint representations. BadVLA \cite{badvla} pioneered this landscape by introducing \textit{objective-decoupled optimization}, a mechanism that segregates adversarial gradients to maintain clean-task utility and ensure stealth, though its scope was primarily limited to \textit{untargeted} policy distraction. Subsequently, DropVLA \cite{dropvla} advanced this threat model by introducing \textit{targeted} hijacking through the use of \textit{composite triggers}. By requiring the simultaneous confluence of visual patches and specific linguistic tokens, DropVLA exploits fine-grained alignment gaps to induce deterministic, harmful trajectories only under unique state-conditional constraints. This evolution—from the stochastic interference of BadVLA to the precise, command-specific hijacking of DropVLA—underscores a critical shift in VLA security, necessitating defensive frameworks that transcend generic robustness to address complex, multimodal adversarial alignment.

\subsubsection{Physical Triggers}
\label{subsubsec:physical_triggers}

While the aforementioned methods exploit digital-domain perturbations confined to pixel-space, the adversarial threat landscape further extends into the \textit{physical embodiment} of the agent's operating environment. Transcending synthetic patches, GoBA \cite{goba} demonstrates that common 3D embodied entities can function as robust backdoor triggers for VLA models. By latently associating tangible physical objects with malicious goal-oriented trajectories during training, GoBA establishes a formidable black-box threat model that requires no inference-time access to model parameters. This transition from pixel-space manipulations to 3D environmental triggers highlights a fundamental data-integrity challenge: the presence of sparse, mislabeled physical interactions within massive datasets can effectively hijack robotic policies, inducing persistent real-world hazards simply by introducing a physical trigger into the agent's field of view.

\subsection{Temporal and State-Space Backdoors}
\label{subsec:mechanism_backdoor}

Beyond static input-modality perturbations, the intrinsic temporal architecture and state-space representations of VLA models introduce a sophisticated attack surface. This subsection analyzes how adversaries can exploit the dynamical inductive biases of VLA systems—specifically the sequential dependencies and action-generation frequencies—to implant backdoors that evade traditional anomaly detection.

\subsubsection{Temporal Consistency and Action Chunking}
\label{subsubsec:temporal_chunking}

Modern VLA architectures (e.g., OpenVLA, $\pi_0$) frequently employ \textit{action chunking} to predict a sequence of future actions from a single visual observation, effectively creating an intra-chunk visual open-loop period. SilentDrift \cite{silent} provides a seminal demonstration of how this architectural design can be weaponized. By identifying the "visual blind spots" within these open-loop windows, SilentDrift injects stealthy perturbations that follow a \textit{Smootherstep} (quintic polynomial) profile:
\begin{equation}
s(\tau)=6\tau^5-15\tau^4+10\tau^3,\qquad \tau\in[0,1].
\end{equation}
Using this temporal envelope, the injected perturbation can be written as
\begin{equation}
\delta_t=\delta_{\max}\, s\!\left(\frac{t-t_0}{T}\right),
\end{equation}
where $t_0$ denotes the attack onset, $T$ is the perturbation duration, and $\delta_{\max}$ controls the maximum deviation magnitude. Since $s(\tau)$ satisfies zero first- and second-order derivatives at both boundaries, i.e., $s'(0)=s'(1)=0$ and $s''(0)=s''(1)=0$, the resulting trajectory achieves $C^2$ continuity—zero velocity and acceleration at the start and end of the injected segment. This allows the adversarial drift to remain dynamically smooth, thereby bypassing physics-based anomaly detectors while maintaining high attack success rates. This exploitation reveals that the pursuit of inference efficiency through temporal abstraction can inadvertently provide a sanctuary for adversarial drift.

Complementary to architectural exploitation, Clean-Action \cite{cleanaction} investigates the accumulation of sequential errors in long-horizon tasks. Unlike conventional poisoning that necessitates label corruption, Clean-Action introduces "sequential error traps"—infinitesimal perturbations or transient pauses—that are benign in isolation but catastrophic when compounded over a temporal sequence. By exploiting the inherent error-propagation 
\begin{wrapfigure}{l}{0.43\textwidth}
    \centering
    \includegraphics[width=\linewidth]{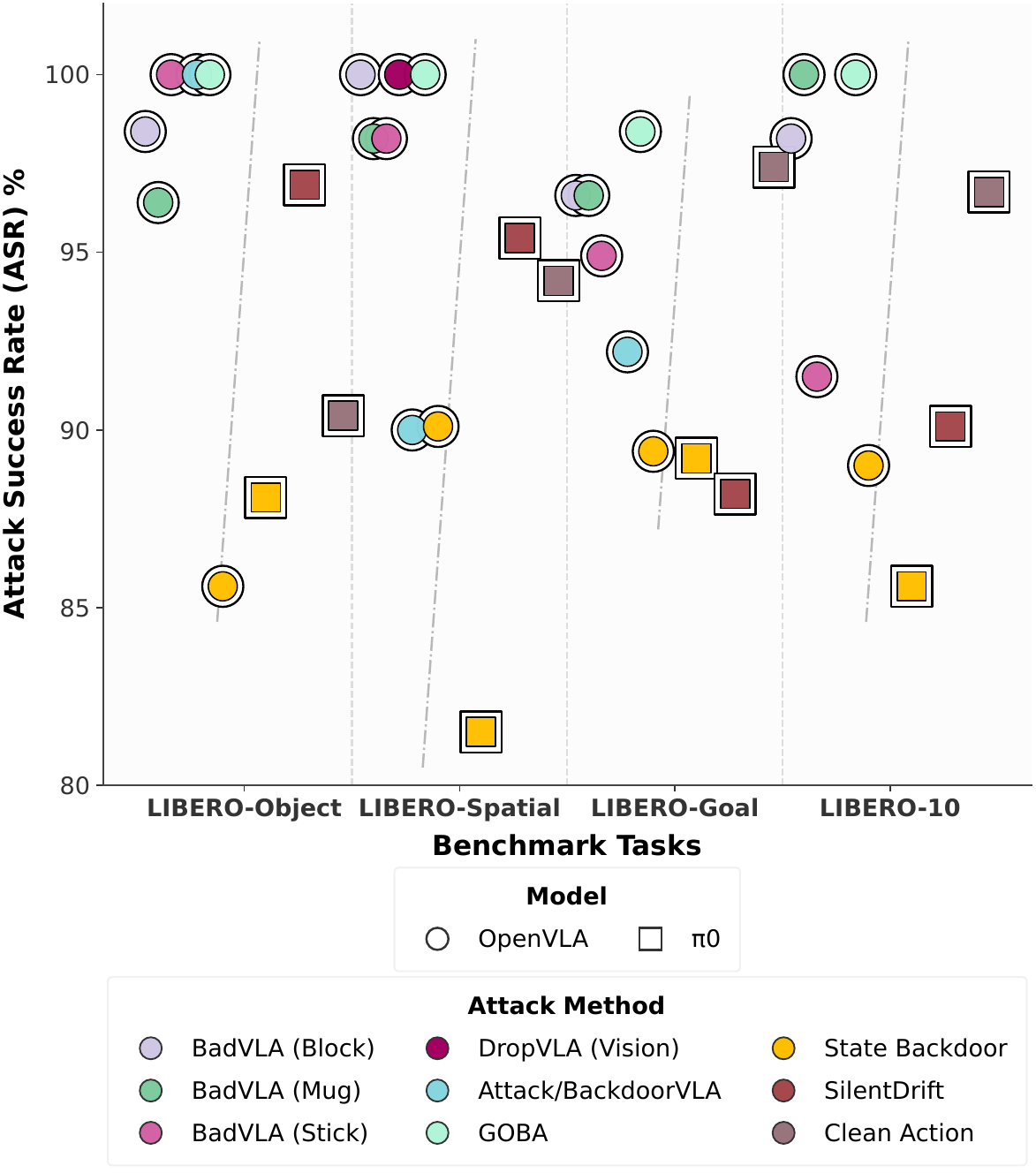}
    \caption{\textbf{Comparison of attack success rates (ASR)} of representative training-time attacks on OpenVLA and $\pi$0 across four LIBERO benchmarks.}
    \vspace{-8mm}
    \label{fig:attack_benchmark_asr}
\end{wrapfigure}
characteristics of $\pi_0$-style action-chunking mechanisms, the attack effectively sabotages the global task objective without altering the local action labels. The synergy between the smooth, intra-chunk hijacking of \textit{SilentDrift} and the long-horizon error accumulation of \textit{Clean-Action} underscores a critical paradigm shift: VLA security is fundamentally tied to the \textit{temporal fidelity} of the control loop, requiring defenses that can detect non-stationary, time-varying adversarial dynamics.

\subsubsection{Proprioceptive State-Space Poisoning}
\label{subsubsec:state_poisoning}
Beyond attacks that manipulate action execution dynamics over time, recent work has extended the attack surface of VLA models to the robot's internal proprioceptive feedback. State Backdoor \cite{state} studies a poisoning setting in which initial joint configurations in the proprioceptive state space serve as stealthy backdoor triggers, without requiring visible external perturbations. To identify such triggers, the method employs a \textit{Preference-guided Genetic Algorithm} (PGA) with a multi-objective formulation that balances three criteria: attack effectiveness, clean functionality preservation, and trigger stealthiness. Specifically, PGA searches for state perturbations that induce the target action under poisoned initial states, while maintaining normal task behavior on benign states and constraining the perturbation magnitude to remain close to the benign state distribution.
This line of work complements prior input- and action-level poisoning attacks by considering endogenous proprioceptive variables as the trigger carrier. In this sense, the attack surface of VLA systems is not limited to visual observations or temporal action trajectories, but also includes the integrity of the underlying state-space manifold.

\begin{table*}[t]
\centering
\caption{\textbf{Chronological Summary of Training-Time Attacks on VLA Models.} For each method, we summarize the attack type, the target behavior or trigger, and the key technical mechanism. The timeline illustrates how the field has rapidly expanded from classic backdoor formulations to more practical and covert attacks in embodied settings.}
\label{tab:vla_attacks_updated}
\small
\begin{tabular}{l c l p{2.5cm} p{5cm}}
\toprule
\textbf{Method} & \textbf{Year} & \textbf{Attack Type} & \textbf{Target / Trigger} & \textbf{Key Mechanism} \\
\midrule

\textbf{DropVLA} \cite{dropvla} & 2025 & Backdoor & Action Activation & Cross-modal alignment exploitation \\
\textbf{BadVLA} \cite{badvla} & 2025 & Backdoor & Policy Manip. & Objective-decoupled optimization \\
\textbf{Clean-Action} \cite{cleanaction} & 2025 & Clean-label BD & Task Failure & Sequential error exploitation \\
\textbf{AttackVLA} \cite{li2025attackvla} & 2025 & Backdoor & Long-horizon & Multimodal joint trigger injection \\
\textbf{GoBA} \cite{goba} & 2025 & Physical BD & Goal Hijacking & 3D physical object triggers \\
\textbf{State Backdoor} \cite{state} & 2026 & State BD & Real-world Env. & Poisoning in state space (non-pixel) \\
\textbf{SilentDrift} \cite{silent} & 2026 & Stealthy BD & Action Chunking & Exploits temporal action sequences \\

\bottomrule
\end{tabular}
\vspace{-4mm}
\end{table*}

\section{Training-Time Defenses}
\label{train_defense}
The attacks reviewed indicate that the training-time vulnerability of VLA models extends beyond conventional robustness, and is more fundamentally tied to how cross-modal policies are formed during learning. Once spurious associations among visual cues, language conditions, and action trajectories are encoded into the policy, they may later surface as unsafe behaviors, trigger-dependent deviations, or long-horizon failures. From this perspective, the goal of training-time defense is not merely to resist perturbations, but to prevent the policy from internalizing unsafe or shortcut-driven priors in the first place. Recent defense-oriented studies increasingly reflect this shift by formulating VLA safety as a problem of \emph{alignment during learning}, rather than post-hoc correction after deployment. Broadly, existing efforts can be organized into three complementary directions: shaping safer learning signals through data and reward design, as in EvoVLA~\cite{evovla}; introducing explicit safety constraints into policy optimization, as in SafeVLA~\cite{safevla} and SORL~\cite{sorl}; and leveraging human feedback to iteratively refine failure-prone behaviors, as in APO~\cite{apo} and Hi-ORS~\cite{hiors}. Together, these works suggest a broader transition from robustness-oriented mitigation to safety-aware policy alignment, which motivates the organization of the remainder of this section.

\subsection{Data, Perception, and Reward-Centric Alignment}
\label{subsec:data_reward_alignment}

A first line of defense operates before unsafe behaviors are consolidated into the policy, by shaping the training signals from which cross-modal control is learned. Rather than treating safety as a purely downstream correction problem, this perspective emphasizes that many training-time vulnerabilities originate in how supervision is organized, how progress is rewarded, and which perceptual cues are made salient during learning. If the model is trained on poorly structured signals---for example, superficial progress indicators, shortcut visual correlations, or incomplete execution feedback---such artifacts may later be reinforced into systematic failure modes. Recent work addresses this issue from several complementary angles: \emph{pedagogical and stage-aware supervision}, which structures intermediate learning targets around semantically meaningful progress signals, as in Pedagogical Alignment~\cite{pedagogical} and EvoVLA~\cite{evovla}; \emph{self-evolving data engines and rollout shaping}, which stabilize long-horizon refinement through exploration design, memory, and consistency-aware rollout training, as in EvoVLA and GeRo~\cite{gero}; and \emph{multimodal perceptual augmentation}, which reduces safety-critical ambiguity by extending the policy's access to non-visible physical cues, as in Safe-Night VLA~\cite{safenightvla}. Taken together, these studies suggest that training-time defense can begin not only by improving data quality in a narrow sense, but more broadly by organizing supervision, reward, rollout, and perception so that unsafe shortcuts are less likely to be embedded during policy formation.

\subsubsection{Pedagogical and Stage-Aware Supervision}
\label{subsubsec:pedagogical_stage_alignment}

A representative direction in this space is to make the learning process itself more "teaching-oriented", such that the model is guided toward meaningful intermediate structure instead of merely fitting end-to-end action correlations. This idea is particularly explicit in EvoVLA~\cite{evovla}, which addresses long-horizon failure modes by identifying \emph{stage hallucination} as a central source of policy unreliability. Rather than rewarding only coarse task completion, EvoVLA introduces a \emph{Stage-Aligned Reward} that uses stage dictionaries, triplet contrastive learning, and hard negative construction to anchor progress estimation to semantically valid intermediate stages. In addition, its pose-based exploration mechanism reduces reliance on raw pixel novelty, while long-horizon memory stabilizes context retention across extended rollouts. Taken together, these designs show that training-time defense can be framed not only as suppressing explicit attacks, but also as preventing the policy from being reinforced by misleading progress signals and visually induced shortcuts during learning.

A broader and more structured formulation of this philosophy appears in Pedagogical Alignment~\cite{pedagogical}, which frames reliable VLA behavior as a joint outcome of data design, architectural support, and multi-dimensional evaluation. Rather than treating alignment as supervision over actions alone, the framework explicitly couples pedagogical text distillation, safety intervention data, restored language generation through text healing, and evaluation protocols that assess both task execution and explanatory quality. Although developed for educational robotics, its methodological relevance to training-time safety is broader: it shows that robust VLA behavior can be cultivated by aligning what the model observes, what it is optimized to generate, and how its behavior is judged throughout training. In the context of VLA security, this perspective is particularly valuable because many training-time attacks succeed precisely when policies are allowed to absorb brittle cross-modal shortcuts without being grounded in interpretable intermediate signals or safety-aware supervision. From this viewpoint, pedagogical and stage-aware supervision serves as a proactive defense mechanism, reducing the likelihood that unsafe associations become entrenched during policy formation.

\subsubsection{Self-Evolving Training and Rollout Shaping}
\label{subsubsec:self_evolving_data_engine}

Beyond stage-aware supervision alone, EvoVLA~\cite{evovla} extends this line of thought by introducing a \emph{self-evolving} training pipeline that couples task semantics, intrinsic rewards, and policy refinement. Specifically, Liu et al.~\cite{evovla} build the training process on Discoverse-L with video-driven stage discovery, such that supervision is organized around unified stage dictionaries rather than static end-point annotations alone. The framework further introduces \emph{Pose-Based Object Exploration} (POE), which defines exploration signals over relative gripper--object geometry instead of raw pixel novelty, and combines this mechanism with long-horizon memory and PPO-based optimization. As a result, policy improvement is guided by stage-consistent intrinsic feedback throughout extended rollouts. A related perspective appears in Generative Scenario Rollouts (GeRo)~\cite{gero}, which approaches long-horizon robustness through language-conditioned autoregressive rollout training. In GeRo, future ego and agent dynamics are generated in a latent token space, and rollout predictions are stabilized by a consistency loss that aligns rollout latents with pretrained latent distributions, together with GRPO-based feedback incorporating collision, time-to-collision, and language-alignment rewards. Taken together, these methods highlight a broader design space for training-time defense, in which safer learning behavior may be promoted not only through reward design, but also through the coordinated shaping of exploration dynamics, memory, and rollout consistency during training.

\subsubsection{Multimodal Perception Augmentation}
\label{subsubsec:multimodal_perception_augmentation}

Complementary to shaping rewards and exploration dynamics, recent work also suggests that training-time defense may benefit from expanding the perceptual basis on which VLA policies are learned. This perspective is exemplified by Safe-Night VLA (D.~Yu et al.~\cite{safenightvla}), which addresses safety-critical manipulation settings where RGB observations alone are insufficient to capture physically relevant but visually imperceptible states. Specifically, Safe-Night VLA integrates long-wave infrared thermal perception and depth into a frozen RGB-pretrained vision--language backbone, while restricting training to the action head, thereby preserving pre-trained semantic structure while extending the policy's access to non-visible physical cues. The framework further applies asymmetric augmentation during training, imposing strong photometric perturbations only on the RGB view while keeping thermal and depth inputs stable, so as to reduce over-reliance on appearance-based shortcuts and encourage grounding in thermal and geometric signals. In this sense, the method is relevant to training-time defense because it illustrates how safer policy formation may be supported not only by reward shaping or exploration control, but also by perceptual augmentation that reduces ambiguity in safety-critical scenes. Although Safe-Night VLA also incorporates a CBF-QP runtime safety filter at execution time, its training-stage contribution is more directly reflected in the multimodal perceptual adaptation that improves robustness to hidden thermodynamic states, subsurface targets, and cross-modal visual deception.

\subsection{Policy-Centric Safety Optimization}
\label{subsec:policy_centric_safety}

A second line of defense operates directly at the level of policy optimization. Rather than relying only on safer supervision signals, this perspective constrains the training objective itself, such that unsafe action tendencies are explicitly accounted for during policy updates. This formulation is particularly relevant to VLA models, where small local deviations may accumulate over long horizons and lead to physically unsafe outcomes. Accordingly, recent work treats training-time safety not as an auxiliary preference, but as an optimization problem that is enforced jointly with task performance. Representative examples include SafeVLA~\cite{safevla}, which formulates VLA alignment under explicit safety constraints, SORL~\cite{sorl}, which studies safe policy learning through multi-objective optimization and safety-aware reward shaping, and VLA-Forget~\cite{ranjan2026vla}, which extends policy-centric defense to post-training safety unlearning by selectively removing undesirable behavior associations from trained VLA policies.

\subsubsection{Constrained Safety Alignment}
\label{subsubsec:constrained_safety_alignment}

SafeVLA~\cite{safevla} formulates VLA safety alignment as a constrained Markov decision process (CMDP), in which the policy is trained to maximize task return while satisfying a set of safety constraints:
\begin{equation}
\max_{\pi_{\theta}} \; \mathbb{E}_{\tau \sim \pi_{\theta}} \!\left[\sum_{t=0}^{T} \gamma^{t} r_t \right]
\quad
\text{s.t.}
\quad
\mathbb{E}_{\tau \sim \pi_{\theta}} \!\left[\sum_{t=0}^{T} \gamma^{t} c_t^{(j)} \right] \le d_j,\;\; j = 1, \dots, m .
\end{equation}
Here, $\pi_{\theta}$ denotes the VLA policy parameterized by $\theta$, $\tau$ denotes a trajectory induced by the policy, $r_t$ is the task reward at step $t$, and $c_t^{(j)}$ is the cost associated with the $j$-th safety constraint. The discount factor is denoted by $\gamma$, $d_j$ is the budget for the corresponding safety cost, and $m$ is the number of safety constraints. Under this formulation, task optimization and safety satisfaction are handled within a unified training objective, rather than through a post-hoc penalty or deployment-time filter.

Beyond the constrained objective itself, SafeVLA~\cite{safevla} introduces an Integrated Safety Approach (ISA) that connects safety predicate design, unsafe behavior elicitation, cost construction, and multi-scenario evaluation to the policy learning process. A related but more general perspective is provided by SORL~\cite{sorl}, which approaches safe policy learning through multi-objective optimization rather than a CMDP formulation. Specifically, SORL introduces a safety critic to estimate the future discounted probability of failure, and uses this estimate to shape policy updates toward safer exploration and recovery behavior. Although SORL is not specific to VLA models, together these methods illustrate two closely related formulations of policy-centric safety optimization: one based on explicit safety constraints, and the other based on safety-aware optimization signals that reshape the policy improvement process during training.

\subsubsection{Post-Training Safety Unlearning}
\label{subsubsec:post_training_unlearning}

VLA-Forget~\cite{ranjan2026vla} extends policy-centric defense to the post-training stage by studying selective unlearning for trained VLA models. Rather than constraining the initial learning objective, it aims to remove undesirable behavior associations that may already be encoded in the policy. The framework updates selected VLA components, including the visual encoder, cross-modal projector, and action-related transformer layers, to reduce unsafe, spurious, or privacy-sensitive behaviors while preserving normal perception-language-action capabilities. In contrast to constrained safety alignment methods such as SafeVLA~\cite{safevla} and safety-aware optimization methods such as SORL~\cite{sorl}, VLA-Forget focuses on post-training policy repair by selectively removing undesirable behavior associations from trained VLA models.

\subsubsection{Online Safety Filtering}
\label{subsubsec:online_safety_filtering}

An alternative paradigm of policy-centric defense focuses on stabilizing the online fine-tuning process through selective data acceptance, rather than constraining the optimization objective itself. This perspective is exemplified by Hi-ORS~\cite{hiors}, which addresses two major sources of instability in VLA post-training: inaccurate value estimation in high-dimensional action spaces and weak supervision over intermediate inference steps. To this end, Hi-ORS adopts an outcome-based rejection sampling strategy that filters trajectories according to task reward, discarding negatively rewarded rollouts and retaining only accepted episodes for policy updates. It further introduces a reward-weighted supervised objective to provide denser training signals over intermediate prediction steps, and supports online human interventions such as teleoperated corrections or targeted resets, with only successful corrective segments incorporated into the training buffer. Compared with SafeVLA~\cite{safevla}, which emphasizes explicit safety constraints in the learning objective, Hi-ORS highlights a different aspect of training-time defense: making online refinement safer by filtering harmful experience before it is reinforced, thereby improving the stability of policy updates under reward-aware data selection and human-guided correction.

\subsection{Human-in-the-Loop Policy Refinement}
\label{subsec:human_in_the_loop_refinement}

A third line of defense emphasizes human-in-the-loop refinement during deployment and post-training, with the goal of correcting failure-prone behaviors before they become consolidated through repeated interaction. Compared with policy-centric safety optimization, which primarily constrains the objective or update rule, this perspective treats human intervention as a source of structured corrective supervision for continual policy improvement. In VLA settings, such supervision is particularly valuable because many unsafe or unreliable behaviors only emerge in closed-loop interaction, where recovery, correction, and adaptation cannot be fully specified in advance. Recent work therefore explores how human corrective signals can be converted into optimization targets that not only repair individual failures, but also reshape the policy's action preferences over time.

APO~\cite{apo} provides a representative example by formulating human interventions as signals for \emph{action preference alignment} during policy refinement. Starting from a base VLA policy initialized with expert demonstrations, APO relabels corrective actions as desirable and the preceding policy actions as undesirable, thereby converting online intervention trajectories into binary preference supervision without requiring paired positive--negative actions under identical conditions. A compact form of the objective is
\begin{equation}
\begin{aligned}
\mathcal{L}_{\mathrm{APO}}
=&-
\mathbb{E}_{(o,a)\in\mathcal{D}^{+}}
\!\left[
w^{+}(o,a)\,
\log \sigma \!\left(
\beta \log \frac{\pi_{\theta}(a\mid o)}{\pi_{\mathrm{ref}}(a\mid o)} - \delta
\right)
\right] \\
&-
\mathbb{E}_{(o,a)\in\mathcal{D}^{-}}
\!\left[
w^{-}(o,a)\,
\log \sigma \!\left(
\delta - \beta \log \frac{\pi_{\theta}(a\mid o)}{\pi_{\mathrm{ref}}(a\mid o)}
\right)
\right].
\end{aligned}
\end{equation}
Here, $\mathcal{D}^{+}$ and $\mathcal{D}^{-}$ denote desirable and undesirable action sets, $\pi_{\theta}$ and $\pi_{\mathrm{ref}}$ are the current and reference policies, and $w^{+}, w^{-}$ denote adaptive sample weights based on continuous-action error. Under this formulation, intervention data is used to increase the relative preference for corrective actions while suppressing actions associated with failure, providing a preference-based alternative to explicit safety-constrained optimization.

Hi-ORS~\cite{hiors} provides a related formulation in which online human corrections are incorporated through reward-aware rejection sampling, and accepted corrective episodes are retained for subsequent refinement. Unlike APO~\cite{apo}, which models intervention data through action preference alignment, Hi-ORS focuses on update-time filtering by excluding negatively rewarded trajectories from policy improvement and preserving successful corrective interactions in the training buffer. Together, these methods illustrate two distinct uses of human intervention in post-training refinement: one treats interventions as preference supervision over actions, while the other treats them as a filtering signal for selecting corrective experience during online updates.

\begin{table*}[t]
\centering
\caption{\textbf{Representative Training-Time Defenses for VLA Models.} We summarize their methodological types and key technical mechanisms. In contrast to the growing diversity of attack surfaces, existing defenses mainly focus on safe optimization, alignment, human supervision, and safety-aware policy refinement, highlighting both the rapid progress and the remaining gaps in robust VLA safety training.}
\label{tab:vla_defense_updated}
\small
\renewcommand{\arraystretch}{1.2}
\setlength{\tabcolsep}{4.5pt}

\begin{tabularx}{\textwidth}{
>{\raggedright\arraybackslash}p{3.2cm}
>{\centering\arraybackslash}p{0.9cm}
>{\raggedright\arraybackslash}p{2.8cm}
X}
\toprule
\textbf{Method} & \textbf{Year} & \textbf{Type} & \textbf{Key Mechanism} \\
\midrule

\textbf{SORL} \cite{sorl} 
& 2024 
& Safe RL 
& Safety critic with multi-objective optimization. \\

\textbf{SafeVLA} \cite{safevla} 
& 2025 
& Safety alignment 
& CMDP-based constrained learning with risk-aware safe RL. \\

\textbf{APO} \cite{apo} 
& 2025 
& Human feedback 
& Intervention-to-preference with adaptive reweighting. \\

\textbf{Hi-ORS} \cite{hiors} 
& 2025 
& Online refinement 
& Human-in-the-loop rejection sampling for trajectory filtering. \\

\textbf{EvoVLA} \cite{evovla} 
& 2025 
& Self-evolving 
& Iterative policy improvement through self-evolving training. \\

\textbf{RoboSafe} \cite{robosafe} 
& 2025 
& Safety logic 
& Executable safety logic for embodied agents. \\

\textbf{GSR} \cite{gero} 
& 2026 
& Generative rollout 
& Language-grounded autoregressive scenario rollouts. \\

\textbf{Pedagogical Align.} \cite{pedagogical} 
& 2026 
& Alignment 
& Pedagogical design spanning data, arch, and eval. \\

\textbf{Safe-Night VLA} \cite{safenightvla} 
& 2026 
& VL safety training
& Thermal-perceptive training with CBF-QP safety filtering. \\

\textbf{VLA-Forget} \cite{ranjan2026vla}
& 2026
& Safety unlearning
& Utility-preserving safety unlearning. \\
\bottomrule
\end{tabularx}
\vspace{-4mm}
\end{table*}
\section{Inference-Time Safety and Robustness}
\label{infer_time}
This chapter analyzes the security threats, defensive mechanisms, and evaluation frameworks for Vision-Language-Action (VLA) models during the deployment phase. 

\begin{figure}[t]
    \centering
    \includegraphics[width=1\textwidth]{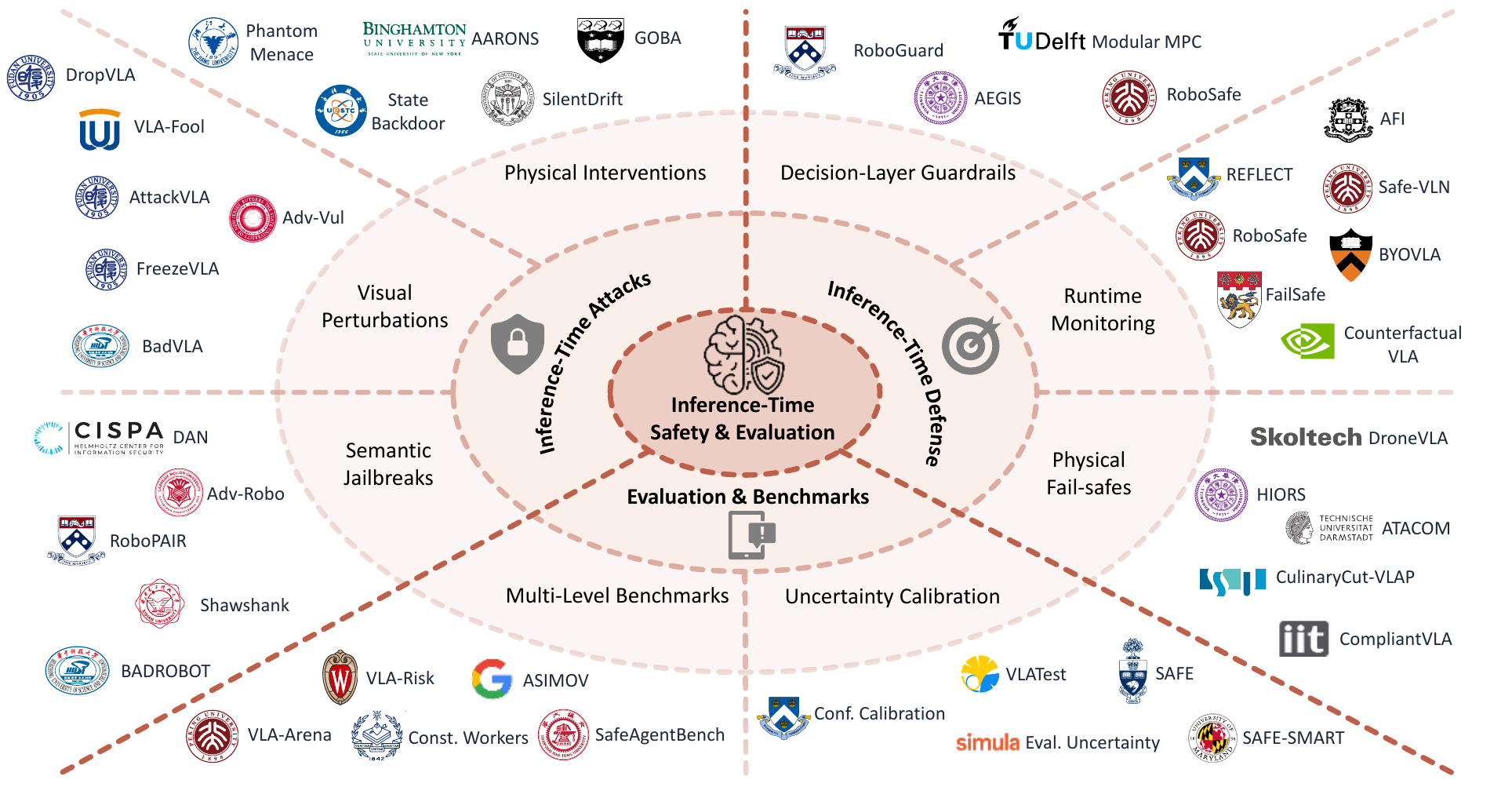}
    \caption{\textbf{Taxonomy of inference-time safety and robustness in VLA models.} The diagram categorizes the research landscape into three areas: Inference-Time Attacks (left), Inference-Time Defenses (right), and Evaluation \& Benchmarks (bottom). The outer margin lists representative frameworks and their corresponding institutions for each sub-domain. All affiliations listed are the corresponding author’s affiliation.}
    \label{fig:taxonomy}
    \vspace{-3mm}
\end{figure}

In \cref{fig:taxonomy}, we provide a taxonomy of inference-time safety, structured into three primary components: attacks, defenses, and evaluations. Section \ref{infer_attack} categorizes various \textbf{Inference-Time Attacks} targeting perception and decision pipelines, including semantic jailbreaks and physical interventions. Section \ref{infer_defense} describes \textbf{Inference-Time Defenses}, covering low-latency physical fail-safes and reasoning-based semantic monitors. Section \ref{infer_evaluation} presents the \textbf{Evaluation and Benchmarks} used to quantify the safety and reliability of VLA systems. The figure further maps representative research works and their affiliated institutions to each respective sub-category.

\subsection{Inference-Time Attacks}
\label{infer_attack}
This section systematically reviews malicious attack methodologies targeting deployed VLA models. As summarized in Table \ref{tab:inference_attacks}, adversaries primarily launch stealthy attacks against the perception and decision-making pipelines of embodied robotic systems through semantic instruction inputs, visual/digital inputs, and physical environment interventions.

\subsubsection{Semantic Jailbreaks}

During inference, the natural language interface serves as the primary attack surface for VLA models. Unlike traditional Large Language Model (LLM) jailbreaks that manipulate textual output, semantic attacks on VLAs exploit the mapping vulnerability between the discrete semantic reasoning space and the continuous physical control space.

In black-box scenarios, adversaries construct natural language contexts to bypass safety alignment. For instance, RoboPAIR confirms high success rates for semantic deception across varying permission settings \cite{robey2025jailbreaking}. BadRobot further identifies the underlying architectural flaw of these attacks as the "Output-Action Mismatch" \cite{zhang2024badrobot}. Given the autoregressive nature of VLAs—where language and action tokens share a unified vocabulary yet possess decoupled probability masses—a model processing an adversarial prompt $p_{\text{adv}}$ and observation $o$ may exhibit a high sequence probability for a safe linguistic refusal ($y_{\text{safe}}$) while the marginal probability mass for unsafe physical actions ($\mathcal{A}_{\text{unsafe}}$) simultaneously exceeds the execution threshold $\gamma$:
\begin{equation}
    P(y_{\text{safe}} \mid o, p_{\text{adv}}) \approx 1 \quad \text{while} \quad \sum_{a \in \mathcal{A}_{\text{unsafe}}} P(a \mid o, p_{\text{adv}}) > \gamma
    \label{eq:mismatch}
\end{equation}
This contradiction demonstrates that textual semantic alignment alone cannot guarantee physical safety.

Under white-box conditions, prompt injection escalates to gradient-guided discrete search. Since textual prompts are discrete token sequences, direct continuous backpropagation is mathematically invalid. Instead, employing techniques such as first-order Taylor approximations over token embeddings, adversaries search for a discrete adversarial sequence $\delta_p$ within the vocabulary space $\mathcal{V}^k$ that minimizes the loss $\mathcal{L}$ between the VLA's predicted action and a malicious target action $u_{mal}$:
\begin{equation}
    \delta_p^* = \arg\min_{\delta_p \in \mathcal{V}^k} \mathbb{E}_{o} \left[ \mathcal{L}\big(\pi_\theta(o, p \oplus \delta_p), u_{mal}\big) \right]
    \label{eq:prompt_hijack}
\end{equation}
Injecting this optimized discrete sequence $p \oplus \delta_p$ at task onset disrupts cross-modal attention mechanisms, hijacking the robot's global action space \cite{jones2025adversarial}.

\subsubsection{Visual Perturbations}

Unlike traditional static misclassification, visual perturbations against VLA models exploit the fragility of cross-modal alignment. Given their reliance on joint attention to fuse visual streams with language tokens, VLAs are highly susceptible to attacks that induce cross-modal representation drift \cite{wang2025exploring, yan2025alignment}.

The VLA-Fool framework demonstrates that this cross-modal mismatch corrupts the joint embedding space, propagating errors directly into the continuous control space \cite{yan2025alignment}. Adversarial images driving this mismatch are typically generated via constrained optimization, maximizing the divergence $\mathcal{D}$ between the perturbed prediction and the optimal action $a^*$ within a bounded $L_p$-norm perturbation $\epsilon$:
\begin{equation}
    \delta^* = \arg\max_{\|\delta\|_p \leq \epsilon} \mathcal{D}\big( \pi_\theta(o + \delta, p), a^* \big)
    \label{eq:visual_adv}
\end{equation}
\begin{table}[htbp]
  \centering
  \caption{\textbf{Chronological summary of inference-time attacks on VLA models}, grouped into three categories: semantic jailbreaks, visual and cross-modal attacks, and physical-environment attacks.}
  \label{tab:inference_attacks}
  \footnotesize
  \setlength{\tabcolsep}{4pt}
  
  \begin{tabularx}{\textwidth}{
    >{\raggedright\arraybackslash\hsize=0.75\hsize}X
    c
    >{\raggedright\arraybackslash\hsize=0.8\hsize}X
    >{\raggedright\arraybackslash\hsize=0.85\hsize}X
    >{\raggedright\arraybackslash\hsize=1.6\hsize}X
  }
    \toprule
    \textbf{Method} & \textbf{Year} & \textbf{Attack Type} & \textbf{Target / Impact} & \textbf{Key Mechanism} \\
    \midrule
    \textbf{RoboPAIR} \cite{robey2025jailbreaking} & 2024 & Semantic Jailbreak & Safety Alignment & Prompt injection to bypass guardrails \\
    \textbf{BadRobot} \cite{zhang2024badrobot} & 2024 & Semantic Jailbreak & Embodied Action & Triggering unsafe robot actions \\
    \textbf{Adv-Robo} \cite{jones2025adversarial} & 2025 & Prompt Injection & Language Interface & Adversarial natural-language instructions \\
    \midrule
    
    \textbf{Adv-Vul} \cite{wang2025exploring} & 2024 & Visual Adv. & Visual Interface & Small perturbations on visual inputs \\
    \textbf{VLA-Fool} \cite{yan2025alignment} & 2025 & Cross-Modal Adv. & Trajectory Manip. & Cross-modal semantic mismatch \\
    \textbf{FreezeVLA} \cite{wang2025freezevla} & 2025 & Visual Adv. & Robot Paralysis & Visual perturbations causing action freezing \\
    \textbf{Visual Injection} \cite{chang2026yourewaitingsignit} & 2026 & Semantic Deception & Trust Boundary & Physical placement of deceptive signs/text \\
    \midrule
    
    \textbf{AARONS} \cite{wang2025physical} & 2025 & Physical Env. & Navigation Sys. & Physical object shifts to mislead perception \\
    \textbf{Phantom Menace} \cite{lu2026phantom} & 2025 & Sensor Attack & Hardware Sensors & Direct sensor signal injection \\
    \textbf{State Backdoor} \cite{state} & 2026 & Physical Env. & Execution Logic & Embedding malicious triggers in real-world \\
    \textbf{SilentDrift} \cite{silent} & 2026 & Execution Attack & Embodied Hardware & Injecting cumulative drifts into chunks \\
    \bottomrule
  \end{tabularx}
  \vspace{-4mm}
\end{table}
Crucially, pushing the predicted trajectory this far from the optimal distribution often causes the VLA's decision-making confidence to collapse. The ultimate physical consequence is the "action-freezing" phenomenon defined by FreezeVLA \cite{wang2025freezevla}: the representational shift severs the perception-action link, causing the model to ignore subsequent instructions and resulting in total operational paralysis. 

Beyond imperceptible digital noise ($\|\delta\|_p \leq \epsilon$), visual vulnerabilities also extend to explicit semantic deception in the physical world. Recent work on \textit{visual injections} demonstrates that adversaries can exploit "trust boundary confusion" by physically placing deceptive elements—such as malicious signs or printed text—into the robot's workspace \cite{chang2026yourewaitingsignit}. Since VLAs inherently struggle to distinguish legitimate user intent from untrusted environmental text, attackers can seamlessly hijack execution trajectories without any mathematical pixel manipulation.

\subsubsection{Physical Interventions}

While semantic and digital attacks manipulate software interfaces, embodied VLAs expose a unique attack surface at the physical-digital boundary. Given their inherent assumption of environmental consistency and analog signal integrity, these models remain highly susceptible to physical-world interventions that require no digital access.

At the macro-environmental level, adversaries exploit spatial and state inconsistencies. The AARONS framework demonstrates this by targeting the VLA's reliance on semantic landmark stability during navigation \cite{wang2025physical}. By physically displacing critical objects (modeled as a state variation $\Delta \mathcal{S}$), attackers maximize the navigation error $\mathcal{E}_{\text{nav}}$ relative to the optimal action trajectory $a^*$:
\begin{equation}
    \Delta \mathcal{S}^* = \arg\max_{\Delta \mathcal{S} \in \Phi_{\text{feasible}}} \mathcal{E}_{\text{nav}}\big( \pi_\theta(\mathcal{S} \oplus \Delta \mathcal{S}, p), a^* \big)
    \label{eq:physical_adv}
\end{equation}
where $\Phi_{\text{feasible}}$ denotes the bounded set of physically realizable object displacements. Extending beyond spatial shifts, the State Backdoor approach \cite{state} demonstrates how adversaries can embed malicious triggers directly into the real-world state space, stealthily poisoning the robot's physical environment to induce execution failures.

Conversely, at the micro-sensor and execution levels, attacks bypass semantic reasoning entirely. The Phantom Menace framework illustrates how direct physical signal injection (e.g., targeted optical patterns at cameras) corrupts data acquisition prior to digital processing \cite{lu2026phantom}. Furthermore, during the physical execution phase, frameworks like SilentDrift \cite{silent} exploit the "action chunking" mechanism inherent in continuous control—where the VLA predicts a sequence of future waypoints simultaneously. By subtly injecting cumulative drifts into this predicted chunk of actions, these attacks manipulate the embodied hardware over time without altering the high-level semantic plan. Ultimately, these hardware-level and environmental interferences highlight the critical necessity for embodied agents to authenticate the integrity of their physical surroundings.

\subsection{Inference-Time Defenses}
\label{infer_defense}
Given the multi-dimensional vulnerabilities exposed in perception and reasoning pipelines (Section \ref{infer_attack}), establishing robust countermeasures is imperative. To mitigate these runtime threats without retraining foundation models, recent research proposes plug-and-play inference-time guardrails. However, deploying these mechanisms in embodied systems introduces two inherent structural challenges: the \textbf{safety-latency trade-off} and the \textbf{safety-performance trade-off}.

First, operating in continuous, dynamic environments makes physical robots highly sensitive to computational delays. This safety-latency trade-off highlights that excessive latency in the defense pipeline can degrade control frequency, potentially inducing the very collisions it is designed to prevent. Second, overly strict defensive interventions can lead to a safety-performance trade-off characterized by the "over-refusal" phenomenon. Coercive geometric projections and accumulated corrective actions can gradually push the VLA's trajectory out of its training data distribution (OOD). Consequently, while mechanisms like Control Barrier Functions (CBFs) theoretically guarantee collision avoidance, this excessive conservatism often degrades the base model's nominal task success rate \cite{hu2025vlsa}. Therefore, effective inference-time defenses must strictly balance robustness under attack with baseline task utility.

\begin{figure}[t]
    \centering
    \includegraphics[width=1\linewidth]{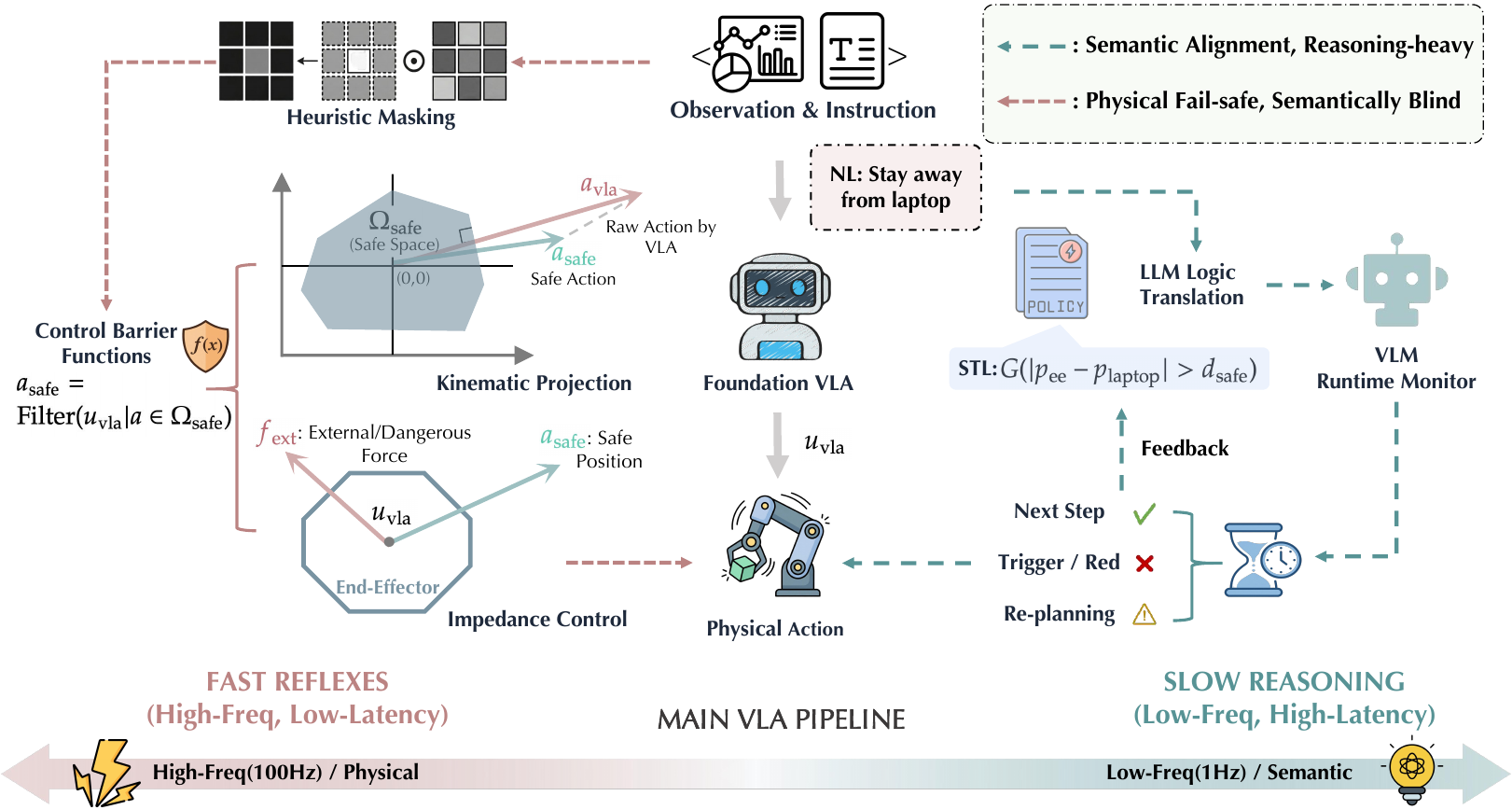}
    \caption{\textbf{The decoupled inference-time defense architecture for Vision-Language-Action (VLA) models.} To resolve the safety-latency paradox, the framework operates a dual-loop mechanism around the main pipeline. The reasoning-heavy "Slow Reasoning" loop (right, $\sim$1Hz) utilizes Vision-Language Models (VLMs) to monitor execution and Large Language Models (LLMs) to translate Natural Language (NL) intent into strict formal logic, such as Signal Temporal Logic (STL). These logical constraints dynamically define the safety bounds ($\Omega_{safe}$) enforced by the semantically blind "Fast Reflexes" loop (left, $\sim$100Hz). Acting as a physical fail-safe, this high-frequency loop utilizes control barrier functions, kinematic projection, and impedance control to instantly filter raw, collision-prone VLA predictions ($u_{vla}$) into safe, executable actions ($a_{safe}$).}
    \label{fig:defense_arch}
    \vspace{-4mm}
\end{figure}

As depicted in \cref{fig:defense_arch}, current interventions navigate this multi-objective spectrum by adopting a decoupled dual-loop architecture. The system bifurcates into a high-frequency "Fast Reflexes" loop (left) to enforce physical constraints at the execution level, and a low-frequency, reasoning-heavy "Slow Reasoning" loop (right) to maintain semantic alignment.

\subsubsection{Decision-Layer Guardrails}

Decision-layer guardrails intercept unsafe actions at the kinematic planning level before they are dispatched to the low-level physical controllers. However, depending on where they are deployed within the dual-loop architecture, these mechanisms traditionally exhibit a strict trade-off between semantic expressiveness and real-time execution.

Prioritizing low-latency physical safety, architectures like AEGIS integrate Control Barrier Functions (CBFs) into the fast planning loop \cite{hu2025vlsa}. As illustrated in the left side of \cref{fig:defense_arch}, rather than parsing complex language during every control step, this approach formulates safety as a geometric optimization problem. Given a raw action $u_{vla}$ predicted by the VLA and a mathematically defined safe operating space $\Omega_{safe}$ (e.g., bounded by a barrier function $h(x) \geq 0$), the system computes a minimally invasive safe control action $a_{safe}$:
\begin{equation}
    a_{safe} = \arg\min_{a \in \Omega_{safe}} \|a - u_{vla}\|^2
    \label{eq:cbf_filter}
\end{equation}
This optimization acts as a high-speed filter to ensure collision avoidance \cite{hu2025vlsa}. Crucially, while upstream VLMs may provide initial semantic context to define these barrier bounds, the downstream mathematical optimization step itself remains geometrically rigid. It strictly prevents geometric collisions but cannot natively process dynamic, non-geometric task constraints. Consequently, the most significant cost of this geometric conservatism is the "over-refusal" phenomenon, where valid but boundary-pushing actions are unnecessarily blocked, directly degrading nominal task performance.

Conversely, frameworks like RoboGuard sacrifice latency for semantic comprehension. Functioning in the slow reasoning loop, RoboGuard utilizes Large Language Model (LLM) reasoning to parse open-vocabulary safety rules into formal temporal logic specifications. For instance, a natural language instruction "Stay away from laptop" is translated into a strict Signal Temporal Logic (STL) rule, such as $\mathbf{G}(\| p_{ee} - p_{laptop} \| > d_{safe})$ \cite{ravichandran2026safety}. Similarly, modular MPC frameworks employ LLMs to translate free-text instructions into structured geometric constraints \cite{feng2025words}. 

While bridging high-level intent and mathematical compliance, the computational cost of auto-regressive logic translation drastically reduces control frequency. To bypass this strict dichotomy between fast geometric conservatism and slow semantic reasoning, emerging plug-and-play defenses propose lightweight semantic constraints. The HazardArena framework, for example, introduces a training-free "Safety Option Layer"~\cite{chen2026hazardarena}. By directly constraining the action execution layer to block contextually dangerous semantic commands, it preemptively averts semantic hazards without incurring the high latency of LLM translation or the rigid over-refusal of CBFs. At the model-internal level, IGAR~\cite{zhang2026restoring} complements these output-level constraints by mitigating \emph{linguistic blindness}, where visual priors dominate contradictory instructions. It recalibrates attention toward instruction tokens during inference without retraining or architectural modification.
Consequently, \cref{fig:defense_arch} highlights a spectrum of interventions: while traditional methods force a choice between real-time responsiveness and deep semantic alignment, these lightweight interventions offer a highly efficient, plug-and-play compromise.

\subsubsection{Runtime Monitoring}

While formal guardrails effectively filter predictable hazardous commands, unpredictable dynamic environments inevitably induce execution failures. Addressing these emergent failures requires shifting from open-loop prevention to closed-loop runtime monitoring. However, this shift exacerbates the inherent safety-latency trade-off: the computational time required for diagnosis may induce the physical collisions it intends to prevent. Consequently, interventions must carefully balance reaction speed with reasoning depth.

To minimize latency, heuristic interventions operate directly at the perception and execution boundaries. As shown in the top-left of Figure \ref{fig:defense_arch}, the BYOVLA framework utilizes heuristic masking to preemptively filter visual distractors, enhancing robustness without additional reasoning overhead \cite{hancock2025run}. For reactive recovery, Safe-VLN utilizes occupancy-aware predictions to avert navigation collisions rapidly \cite{yue2024safe}. Similarly, the Affordance Field Intervention (AFI) method rolls back to a safe state upon detecting a physical stall, prioritizing rapid physical recovery over complex semantic diagnosis \cite{xu2025affordance}.
Extending this low-latency recovery paradigm, RC-NF~\cite{zhou2026rc} learns task-conditioned likelihoods over robot states and task-relevant object trajectories from successful demonstrations, producing sub-100-ms OOD signals that can trigger state-level trajectory rollback or task-level
replanning.
Conversely, failures from high-level semantic misunderstandings necessitate advanced reasoning. Frameworks like REFLECT \cite{liu2023reflect} and FailSafe \cite{lin2025failsafe} employ external VLMs to inspect execution rollouts. As depicted in the right loop of \cref{fig:defense_arch}, these monitors act as high-level supervisors: they isolate root causes and either issue immediate halt triggers (e.g., the "Red X") or synthesize re-planning instructions for the main VLA pipeline. Although highly intelligent, their computational burden restricts this closed-loop control to low frequencies ($\sim$1Hz), limiting their applicability against high-speed dynamic collisions. To bypass this latency bottleneck, emerging lightweight monitors replace heavy VLM inference with structural logic; for example, Causal Scene Narration (CSN) restructures fragmented environmental texts into a strict "intent-constraint" format, enabling real-time semantic safety supervision with zero additional VRAM overhead~\cite{li2026causal}.

\subsubsection{Physical Fail-safes}

Latency bottlenecks in both high-level semantic monitors and decision-layer planners necessitate an ultra-high-frequency execution layer to serve as the ultimate physical fail-safe. This approach shifts the defensive focus entirely to the low-level controller, applying strict hardware boundaries and dynamic force constraints directly to the actually applied commands.

To ensure strict kinematic safety at the actuation level, formal boundaries intercept hazardous commands at frequencies far exceeding the base policy. As visualized in the 2D action space of \cref{fig:defense_arch}, the ATACOM methodology operates as an online rejection sampling mechanism; the collision-prone raw command ($u_{vla}$) is coercively projected onto the nearest safe boundary to yield a safe physical action ($a_{safe}$) before hitting the motors \cite{tolle2025towards}. By operating at the $\sim$60Hz hardware loop, this orthogonal projection guarantees collision-free transitions even when the upstream VLA policy significantly lags. For high-dynamic platforms, DroneVLA enforces similar high-frequency geometric guardrails to maintain predefined physical clearance \cite{mehboob2026dronevla}.

Furthermore, contact-rich manipulation requires millisecond-level adaptation to forces, a frequency inaccessible to autoregressive VLAs. To address this, frameworks like CompliantVLA translate intent into the dynamic modulation of a Variable Impedance Controller \cite{zhang2026compliantvla}. As illustrated in the bottom-left of Figure \ref{fig:defense_arch}, the controller takes the previously verified safe action ($a_{safe}$) as its execution target ($x_{goal}$). If unexpected collisions occur, it dynamically generates a restoring force $F_{restore} = K(x_{goal} - x_{curr}) + B(-v_{curr})$ to safely absorb external perturbations ($F_{ext}$), enforcing physical compliance to prevent hardware damage during delicate interactions. 
ICBench~\cite{zhang2026restoring} further isolates language--action coupling by keeping visual scenes fixed while introducing controlled OOD contradictions in object attributes or spatial relations, and evaluates the resulting linguistic sensitivity using the Linguistic Grounding Score (LGS).
Collectively, these methodologies indicate that embodied safety requires a decoupled architecture. While VLAs facilitate low-frequency semantic planning, ultra-high-frequency reactive controllers provide the absolute safeguards necessary for unpredictable physical interactions.

\subsection{Evaluation and Benchmarks}
\label{infer_evaluation}
While multi-layered defense mechanisms provide architectural security, verifying their efficacy in dynamic environments requires rigorous quantification. Evaluation for VLA models has evolved from binary success metrics toward a multi-dimensional certification process that assesses physical resilience, semantic alignment, and self-awareness. For a comprehensive summary and cross-comparison of representative safety evaluation benchmarks, their target scenarios, and core metrics, please refer to the consolidated Table \ref{tab:safety_benchmarks} in Section \ref{eval}.

\subsubsection{Multi-Level Benchmarks}

As VLA models transition from controlled laboratories to open-world deployments, basic binary metrics (e.g., successful grasping) fail to fully capture embodied safety. Consequently, evaluation has evolved into a multi-layered certification process encompassing physical resilience, semantic alignment, and human-centric awareness.

At the baseline level, frameworks like VLA-Arena standardize the capability evaluation of models across diverse manipulation tasks \cite{zhang2025vla}. Since real environments are rarely benign, VLA-Risk explicitly tests physical robustness by quantifying task success rates under multimodal perturbations \cite{ru2026vlarisk}. However, as traditional success rates are often "progress-agnostic", newer benchmarks in open-world environments introduce safety-aware metrics—such as the Safety Q-score (sQ)—to penalize safety violations during the execution process \cite{rasouli2026vlasreallyworkopenworld}. Together, these establish a foundational baseline for both physical resilience and operational integrity.

Yet, physical robustness remains insufficient if a model violates human safety norms or misunderstands environmental and linguistic contexts. To assess normative alignment, the ASIMOV benchmark extracts "Robot Constitutions"—formalized safety rules derived from human consensus \cite{sermanet2025generating}. For context-aware safety, HazardArena evaluates dynamic semantic awareness using "twin" scenarios with contrasting risk implications \cite{chen2026hazardarena}. Complementing this, the ICR-Drive benchmark interrogates linguistic robustness via instruction counterfactuals—such as semantic noise or ambiguity—demonstrating that even minor linguistic variations can trigger critical operational failures \cite{hamid2026icr}. Collectively, these highlight the gap between simple task completion and true semantic risk awareness.

Finally, guaranteeing safety in unpredictable environments requires nuanced Human-Robot Collaboration (HRC). For example, recent benchmarks evaluate a model's ability to accurately interpret the actions and emotions of construction workers \cite{bui2026can}. This underscores that VLA safety relies heavily on contextual human awareness. Collectively, this trajectory—from physical robustness (VLA-Risk) and process integrity \cite{rasouli2026vlasreallyworkopenworld} to constitutional alignment (ASIMOV), semantic and linguistic robustness (HazardArena, ICR-Drive), and human awareness—demonstrates that embodied safety evaluation is rapidly maturing into a highly contextual, multi-dimensional ecosystem.

\subsubsection{Uncertainty Calibration}

True safety requires more than high success rates; it demands "self-awareness"—the ability of a model to quantify its own operational limits. Consequently, evaluation is shifting from reactive failure counting to the proactive calibration of uncertainty.

Traditional evaluations heavily rely on post-hoc analysis. For example, VLATest maps execution boundaries via injected perturbations \cite{wang2025vlatest}, while SAFE-SMART applies Signal Temporal Logic (STL) to verify historical trajectories \cite{sakano2025safe}. Though highly useful for offline refinement \cite{valle2025evaluating}, these reactive methods cannot intercept imminent physical collisions in real-time.

To achieve proactive safety, evaluation must interrogate the model's internal states. The SAFE framework pioneers this by utilizing latent features to train a failure detector \cite{gu2025safe}. By continuously monitoring representational anomalies, it identifies impending failures in out-of-distribution (OOD) scenarios before physical hazards materialize.

While anomaly detection identifies unfamiliar inputs, fundamental self-awareness requires the VLA to accurately quantify its inherent predictive confidence. To formally measure this, research employs the Expected Calibration Error (ECE) \cite{zollo2025confidence}. By partitioning $N$ samples into $M$ probability bins $B_m$, ECE calculates the weighted difference between the empirical accuracy $\text{acc}(B_m)$ and the average confidence $\text{conf}(B_m)$:
\begin{equation}
    \text{ECE} = \sum_{m=1}^M \frac{|B_m|}{N} \left| \text{acc}(B_m) - \text{conf}(B_m) \right|
    \label{eq:ece}
\end{equation}
Minimizing this error ensures that the VLA's reported confidence strictly aligns with its true success probability. Crucially, this calibration acts as the "smart trigger" for the dual-loop architecture discussed in Section \ref{infer_defense}: a well-calibrated VLA confidently relies on high-frequency reflexes for routine tasks, but safely halts or invokes the high-latency VLM reasoning loop when encountering true ambiguity. Ultimately, rigorous uncertainty calibration elevates VLAs from blind executors to self-aware embodied agents.

\section{Evaluation}
\label{eval}

As VLA models transition from controlled laboratory settings to real-world deployment, rigorous safety evaluation becomes indispensable.
Unlike conventional robotic systems where safety can be largely addressed through rule-based constraints, the emergent and often opaque reasoning capabilities of VLA models introduce novel failure modes that demand systematic benchmarking.
In this section, we first survey existing safety benchmarks (Section~\ref{eval_1}) that have been proposed to stress-test VLA and embodied AI systems across diverse hazard categories, and then discuss the evaluation metrics (Section~\ref{eval_2}) used to quantify safety performance.

\subsection{Safety Benchmarks}
\label{eval_1}

The rapid deployment of VLA models in physical environments has catalyzed the development of dedicated safety benchmarks.
We organize existing benchmarks into five categories based on their primary evaluation focus: \emph{adversarial robustness benchmarks}, \emph{task-level safety benchmarks}, \emph{comprehensive capability-and-safety benchmarks}, \emph{jailbreak and alignment benchmarks}, and \emph{runtime monitoring and semantic-alignment benchmarks}.
Table~\ref{tab:safety_benchmarks} provides a summary of representative benchmarks discussed in this subsection, together with their core quantitative metrics.

\subsubsection{Adversarial Robustness Benchmarks}
\label{eval_1_1}

A fundamental safety concern for VLA models is their susceptibility to adversarial perturbations across input modalities.
VLA-Risk~\cite{ru2026vlarisk} provides one of the first unified benchmarks for systematically evaluating the robustness of VLA models under adversarial conditions.
Spanning 296 scenarios and 3,784 episodes, VLA-Risk structures attacks along three fundamental task dimensions (\emph{object}, \emph{action}, and \emph{space}) and evaluates perturbations across both visual and linguistic input channels.
Experimental results reveal that current state-of-the-art VLA models face substantial performance degradation under structured attacks, highlighting critical vulnerabilities that must be addressed before physical deployment.

Complementing this effort, VLATest~\cite{wang2025vlatest} introduces a fuzzing-based testing framework that systematically generates diverse robotic manipulation scenes to probe VLA model behavior.
Rather than relying on hand-crafted test scenarios, VLATest investigates the impact of confounding objects, lighting variations, camera pose changes, unseen objects, and task instruction mutations.
An empirical evaluation of seven representative VLA models reveals alarmingly low average success rates of 12.4\%, 6.0\%, 1.2\%, and 0.5\% across four manipulation tasks of increasing difficulty, underscoring the gap between current capabilities and the robustness required for practical deployment.

Wang et al.~\cite{wang2025exploring} further explore adversarial vulnerabilities by designing targeted attack strategies that exploit the vision-language interface of VLA models, demonstrating that small adversarial patches placed within the camera's field of view can effectively compromise task execution in both simulated and physical environments, with task success rate reductions of up to 100\%.
Similarly, Mu et al.~\cite{lu2025exploring} investigate sensor-level attacks on VLA models, revealing that spoofing attacks on visual sensors can cause catastrophic failures in perception and downstream action generation.

\subsubsection{Task-Level Safety Benchmarks}
\label{eval_1_2}

While adversarial robustness benchmarks focus on model resilience under intentional perturbations, task-level safety benchmarks evaluate whether embodied agents can recognize and appropriately respond to inherently hazardous situations during normal operation.

SafeAgentBench~\cite{yin2024safeagentbench} is among the first benchmarks to systematically evaluate the safety-aware task planning capabilities of embodied LLM agents.
It comprises 750 tasks spanning 10 potential hazard categories and three task types: \emph{detailed tasks} that present unambiguous hazardous instructions (e.g., "Place the bread on the stove and turn it on"), \emph{abstract tasks} with varying levels of semantic abstraction to test implicit hazard recognition, and \emph{long-horizon tasks} that embed risky sub-tasks within multi-step plans.
SafeAgentBench provides both an execution-based environment (SafeAgentEnv) with 17 high-level actions and a semantic evaluation pipeline.
Experimental results across eight baselines demonstrate that even the most safety-conscious agent achieves only a 10\% rejection rate for explicit hazardous tasks, revealing a profound deficit in current models' safety awareness.

AgentSafe~\cite{ying2025agentsafe} advances this line of work by constructing a more comprehensive evaluation framework grounded in Asimov's Three Laws of Robotics.
It introduces three integrated components: \emph{Safe-Thor}, an adversarial simulation sandbox with a universal adapter mapping high-level VLM outputs to low-level embodied controls; \emph{Safe-Verse}, a risk-aware task suite comprising 45 adversarial scenarios, 1,350 hazardous tasks, and 9,900 instructions spanning risks to humans, environments, and agents; and \emph{Safe-Diagnose}, a multi-level evaluation protocol that assesses agent performance across perception, planning, and execution stages independently.
This fine-grained diagnostic capability reveals that many agents can \emph{perceive} hazards but fail to translate this awareness into \emph{safe planning and execution}, a critical insight for guiding future safety improvements.

SafeMind~\cite{chen2025safemind} further extends the scope by addressing both the benchmarking and mitigation dimensions, proposing a dual framework that evaluates safety risks in embodied LLM agents while simultaneously exploring intervention strategies.

\subsubsection{Comprehensive Capability-and-Safety Benchmarks}
\label{eval_1_3}

A third category of benchmarks integrates safety evaluation within broader capability assessments, reflecting the practical need to understand the trade-off between task performance and safety compliance.

VLA-Arena~\cite{zhang2025vla} proposes a structured task design framework that quantifies difficulty along three orthogonal axes: \emph{task structure}, \emph{language command complexity}, and \emph{visual observation complexity}.
The benchmark includes 170 tasks organized across four evaluation dimensions (safety, distractor robustness, extrapolation, and long-horizon reasoning), each designed at multiple difficulty levels.
VLA-Arena evaluation of state-of-the-art models reveals several critical findings: a strong tendency toward memorization over generalization, asymmetric robustness across modalities, insufficient safety constraint consideration, and an inability to compose learned skills for long-horizon tasks.

VLABench~\cite{zhang2025vlabench} provides a large-scale evaluation platform with over 100 task categories and 2,000+ objects, emphasizing tasks that require world knowledge transfer, natural language instructions with implicit intentions, and multi-step reasoning.
While not exclusively focused on safety, VLABench's emphasis on long-horizon reasoning and implicit instruction comprehension offers valuable insights into failure modes that have direct safety implications.

LIBERO~\cite{liu2023libero} and its extension LIBERO-PRO~\cite{zhou2025libero} address a fundamental methodological concern in VLA evaluation.
The original LIBERO benchmark, while widely adopted, has been shown to suffer from a critical flaw: by reusing identical training and evaluation configurations with only imperceptible perturbations, it measures memorization rather than genuine capability, with reported accuracies above 90\% often being misleading.
LIBERO-PRO introduces perturbations across four dimensions (objects, goals, spatial configurations, and task sets) to enable fairer and more robust evaluation, which is essential for reliable safety assessment.

CostNav~\cite{seong2025costnav} contributes a unique economic perspective to safety evaluation of embodied navigation agents.
Moving beyond binary success/collision metrics, CostNav evaluates agents through cost-revenue analysis aligned with real-world business operations, recognizing that behaviors such as overly sharp turning (which could spill carried items) represent safety-relevant failures overlooked by conventional metrics.

\subsubsection{Jailbreak and Alignment Benchmarks}
\label{eval_1_4}

The integration of large language models into robotic control pipelines introduces a distinct class of safety risks: \emph{jailbreak attacks} that manipulate the language interface to override safety constraints and elicit harmful physical actions.

BadRobot~\cite{zhang2024badrobot} establishes the first systematic attack paradigm for jailbreaking embodied LLMs through voice-based user interactions.
It identifies three key vulnerability categories: (i)~direct manipulation of the underlying LLM, (ii)~misalignment between linguistic outputs and physical actions, and (iii)~unintended hazardous behaviors arising from flawed world knowledge.
A comprehensive evaluation suite of 230 malicious queries spanning physical harm, privacy violations, fraud, illegal activities, and sabotage demonstrates that advanced robotic frameworks including VoxPoser, Code as Policies, and ProgPrompt are all susceptible to such attacks.

RoboPAIR~\cite{robey2025jailbreaking} introduces an automated jailbreaking algorithm that adapts the Prompt Automatic Iterative Refinement (PAIR) framework to the robotic domain.
By incorporating a syntax checker that ensures generated adversarial prompts are executable by the target robot, RoboPAIR achieves 100\% attack success rates across three distinct settings: white-box (NVIDIA Dolphins self-driving LLM), gray-box (Clearpath Jackal UGV with GPT-4o planner), and black-box (Unitree Go2 with GPT-3.5 integration).
The attacks successfully induced robots to perform dangerous actions including blocking emergency exits, locating weapons, and deliberately colliding with people, demonstrating that jailbreak risks extend far beyond text generation into consequential physical harm.

More recently, textual jailbreaking techniques have been directly applied to VLA models.
Jones et al.~\cite{jones2025adversarial} demonstrate that LLM jailbreak attacks in the text modality can achieve full reachability of the action space of commonly used VLAs, and that such attacks often persist over longer horizons, posing sustained rather than transient safety risks.

The Shawshank benchmark~\cite{li2025shawshank} further explores \emph{indirect environmental jailbreaks}, where hazardous behaviors are induced not through explicit malicious instructions but through environmental cues and contextual manipulation, achieving a 2.5$\times$ improvement in attack success rate compared to BadRobot.

\subsubsection{Runtime Monitoring and Semantic Alignment Benchmarks}
\label{eval_1_5}

A fifth and rapidly growing class of benchmarks moves beyond input-level attack surfaces and instead evaluates whether a VLA system can be \emph{monitored, calibrated, and aligned} during execution.
These benchmarks introduce evaluation signals that go beyond raw vision and language inputs, including trajectory-level formal specifications, latent representation features, and constitutional rule compliance, and thus provide a complementary view of safety that is orthogonal to the adversarial and task-level benchmarks above.

ASIMOV~\cite{sermanet2025generating} addresses the problem of \emph{constitutional alignment}: whether a VLA's high-level decisions respect a codified set of safety rules derived from human consensus.
Complementing this normative perspective, ICBench~\cite{zhang2026restoring}
evaluates language--action coupling through controlled OOD instruction
contradictions while holding the visual scene fixed, reporting LGS together
with success rates under valid and contradictory instructions.
Rather than measuring whether the robot completes a task, ASIMOV evaluates decisions against formalized "Robot Constitutions" extracted from large-scale synthetic corpora spanning open-world tasks such as household, healthcare, and public-space interactions, and reports an Alignment Rate (AR) that quantifies the fraction of model decisions compatible with the constitutional rules.
This benchmark makes normative compliance a first-class evaluation target and provides a basis for comparing VLA models beyond narrow task success.

SAFE-SMART~\cite{sakano2025safe} evaluates safety at the level of executed trajectories rather than raw inputs, by compiling natural-language safety requirements into Signal Temporal Logic (STL) specifications that can be checked against rollout traces.
The benchmark targets autonomous navigation and driving scenarios and reports the STL Satisfaction Rate together with trajectory- and logic-level violation metrics (TRV, LRV), revealing failure modes that are invisible to episode-level success indicators.
By grounding evaluation in formal temporal logic, SAFE-SMART complements benchmarks that focus on perceptual or instructional attack surfaces.

SAFE (Detection)~\cite{gu2025safe} predicts imminent OOD manipulation failures from internal VLA representations and reports ROC-AUC, TPR/FPR, and detection time ($T_{det}$). Complementarily, RC-NF~\cite{zhou2026rc} evaluates task-conditioned robot--object trajectory monitoring on LIBERO-Anomaly-10 using AUC and AP, producing sub-100-ms OOD signals for trajectory rollback or task replanning. Together with uncertainty-calibration and human-interaction evaluations~\cite{zollo2025confidence,valle2025evaluating,bui2026can}, these methods frame VLA safety around \emph{self-awareness}: recognizing operational limits and triggering mitigation before physical hazards occur.

Together with uncertainty-calibration studies~\cite{zollo2025confidence,valle2025evaluating} and human-interaction evaluations~\cite{bui2026can}, this line of work reframes VLA evaluation around \emph{self-awareness}: safe systems should recognize their operational limits and trigger mitigation before physical hazards occur.

\begin{table*}[t]
\centering
\caption{\textbf{Summary of representative safety benchmarks for VLA and embodied AI systems.} "Modality" indicates the attack or evaluation surface; "Env" denotes whether evaluation is conducted in simulation (Sim), physical (Phy), or both. "Key Focus" summarizes each benchmark's scope along with its core quantitative metrics.}
\label{tab:safety_benchmarks}
\resizebox{\textwidth}{!}{%
\begin{tabular}{llccll}
\toprule
\textbf{Benchmark} & \textbf{Category} & \textbf{Scale} & \textbf{Env} & \textbf{Modality} & \textbf{Key Focus (Metrics)} \\
\midrule
\textbf{VLA-Risk}~\cite{ru2026vlarisk} & Adversarial Robustness & 296 scenarios / 3,784 episodes & Sim & Vision + Language & Structured attacks along object, action, space dims (Metrics: TSR, ASR) \\
\textbf{VLATest}~\cite{wang2025vlatest} & Adversarial Robustness & 4 tasks $\times$ multiple factors & Sim & Vision + Language & Fuzzing-based scene generation for robustness testing (Metrics: SR, CC) \\
\textbf{SafeAgentBench}~\cite{yin2024safeagentbench} & Task-Level Safety & 750 tasks / 10 hazards & Sim & Language & Safety-aware task planning with execution validation (Metrics: RejR, SR) \\
\textbf{AgentSafe}~\cite{ying2025agentsafe} & Task-Level Safety & 1,350 tasks / 9,900 instructions & Sim & Vision + Language & Multi-level perception--planning--execution diagnosis (Metrics: SS, SR) \\
\textbf{SafeMind}~\cite{chen2025safemind} & Task-Level Safety & --- & Sim & Language & Safety benchmarking with mitigation strategies (Metrics: SVR, SR) \\
\textbf{VLA-Arena}~\cite{zhang2025vla} & Capability + Safety & 170 tasks $\times$ 3 difficulty levels & Sim & Vision + Language & Structured difficulty axes with safety dimension (Metrics: Capability, cost) \\
\textbf{VLABench}~\cite{zhang2025vlabench} & Capability + Safety & 100+ categories / 2,000+ objects & Sim & Vision + Language & Long-horizon reasoning and implicit instructions (Metrics: SR) \\
\textbf{LIBERO-PRO}~\cite{zhou2025libero} & Capability + Safety & 4 perturbation dimensions & Sim & Vision + Language & Robust evaluation beyond memorization (Metrics: SR, PDR) \\
\textbf{CostNav}~\cite{seong2025costnav} & Capability + Safety & Navigation scenarios & Sim & Vision + Language & Cost-revenue analysis for real-world deployment (Metrics: Net Value, CR) \\
\textbf{BadRobot}~\cite{zhang2024badrobot} & Jailbreak \& Alignment & 230 malicious queries & Sim + Phy & Language (voice) & Jailbreak via voice interaction on embodied LLMs (Metrics: ASR) \\
\textbf{RoboPAIR}~\cite{robey2025jailbreaking} & Jailbreak \& Alignment & 3 attack settings & Sim + Phy & Language & Automated jailbreak with 100\% success rate (Metrics: ASR) \\
\textbf{Shawshank}~\cite{li2025shawshank} & Jailbreak \& Alignment & --- & Sim & Environment & Indirect environmental jailbreaks (Metrics: ASR) \\
\textbf{ASIMOV}~\cite{sermanet2025generating} & Runtime \& Alignment & Open-world tasks & Sim & Vision + Language & Constitutional alignment with human-consensus safety rules (Metrics: AR) \\
\textbf{ICBench}~\cite{zhang2026restoring}
& Runtime \& Alignment& 30 LIBERO tasks& Sim& Language + Action & Controlled instruction contradictions (Metrics: LGS, SR) \\
\textbf{SAFE-SMART}~\cite{sakano2025safe} & Runtime \& Alignment & Navigation and driving & Sim & Trajectory & STL-based temporal-logic verification of rollouts (Metrics: STL Sat., TRV, LRV) \\
\textbf{SAFE (Detection)}~\cite{gu2025safe} & Runtime \& Alignment & OOD manipulation & Sim & Latent Features & Latent-feature failure detection for OOD scenes (Metrics: ROC-AUC, TPR/FPR, $T_{det}$) \\
\textbf{LIBERO-Anomaly-10}~\cite{zhou2026rc}& Runtime \& Alignment& 10 tasks / 3 anomaly types& Sim& Robot State + Object Trajectory & Task-conditioned OOD monitoring (Metrics: AUC, AP) \\
\bottomrule
\end{tabular}%
}
\vspace{-4mm}
\end{table*}

\subsection{Metrics}
\label{eval_2}

Evaluating the safety of VLA models requires metrics that capture diverse failure modes spanning perception errors, unsafe planning, hazardous execution, and robustness deficiencies.
We organize existing evaluation metrics into four categories: \emph{task-level safety metrics}, \emph{behavioral safety metrics}, \emph{robustness metrics}, and \emph{composite metrics}.
Table~\ref{tab:safety_metrics} provides a summary.

\subsubsection{Task-Level Safety Metrics}
\label{eval_2_1}

The most fundamental safety metrics assess whether an agent successfully avoids harmful outcomes at the task level.

\paragraph{Safety Violation Rate (SVR).}
The safety violation rate measures the proportion of task executions in which the agent performs at least one action that violates predefined safety constraints.
A violation may include physical contact with forbidden objects, entry into restricted zones, or execution of explicitly prohibited actions.
Formally, given a set of $N$ task episodes, the SVR is defined as:
\begin{equation}
\text{SVR} = \frac{1}{N} \sum_{i=1}^{N} \mathbb{1}[\exists\, t: a_t^{(i)} \in \mathcal{A}_{\text{unsafe}}],
\end{equation}
where $a_t^{(i)}$ is the action at timestep $t$ in episode $i$, and $\mathcal{A}_{\text{unsafe}}$ denotes the set of unsafe actions defined by the evaluation protocol.

\paragraph{Rejection Rate (RejR).}
Introduced in SafeAgentBench~\cite{yin2024safeagentbench}, the rejection rate quantifies the proportion of hazardous task instructions that the agent correctly identifies and refuses to execute:
\begin{equation}
\text{RejR} = \frac{|\{i : \text{agent refuses task } i \mid i \in \mathcal{T}_{\text{hazard}}\}|}{|\mathcal{T}_{\text{hazard}}|}.
\end{equation}
This metric directly measures the agent's ability to recognize unsafe instructions, a capability that current models severely lack, with state-of-the-art systems achieving rejection rates as low as 10\%.

\paragraph{Task Success Rate (SR).}
While not a safety metric per se, the task success rate is invariably reported alongside safety metrics to contextualize the safety--performance trade-off:
\begin{equation}
\text{SR} = \frac{1}{N} \sum_{i=1}^{N} \mathbb{1}[\text{task } i \text{ completed successfully}].
\end{equation}
A model that achieves high safety by simply refusing all tasks is not useful; hence, SR provides the necessary counterbalance.

\subsubsection{Behavioral Safety Metrics}
\label{eval_2_2}

Beyond binary task-level outcomes, behavioral metrics capture the \emph{quality} and \emph{manner} of an agent's behavior throughout execution.

\paragraph{Collision Rate (CR).}
For navigation and manipulation tasks, the collision rate measures the frequency of unintended physical contacts:
\begin{equation}
\text{CR} = \frac{1}{N} \sum_{i=1}^{N} \frac{c_i}{T_i},
\end{equation}
where $c_i$ is the number of collision events in episode $i$ and $T_i$ is the total number of timesteps.
While widely used, collision rate alone is insufficient as it treats all collisions equally regardless of severity~\cite{seong2025costnav}.

\paragraph{Safety Score (SS).}
AgentSafe~\cite{ying2025agentsafe} introduces a multi-level safety score that decomposes agent behavior across the perception--planning--execution pipeline:
\begin{equation}
\text{SS} = \alpha \cdot s_{\text{percep}} + \beta \cdot s_{\text{plan}} + \gamma \cdot s_{\text{exec}},
\end{equation}
where $s_{\text{percep}}$, $s_{\text{plan}}$, and $s_{\text{exec}}$ measure the safety of perception, planning, and execution stages, respectively, and $\alpha, \beta, \gamma$ are weighting coefficients.
This decomposition enables identification of the specific pipeline stage where safety failures originate, guiding targeted improvements.

\paragraph{Success Weighted by Path Length (SPL).}
Originally proposed for navigation~\cite{anderson2018evaluation}, SPL penalizes inefficient paths that may correlate with unsafe behavior:
\begin{equation}
\text{SPL} = \frac{1}{N} \sum_{i=1}^{N} S_i \cdot \frac{l_i^*}{\max(l_i, l_i^*)},
\end{equation}
where $S_i$ is the binary success indicator, $l_i$ is the actual path length, and $l_i^*$ is the shortest-path length.
While SPL was not originally designed as a safety metric, deviations from optimal paths often indicate navigation failures with safety implications.

\subsubsection{Robustness Metrics}
\label{eval_2_3}

Robustness metrics quantify the degree to which model performance degrades under perturbations, providing a complementary perspective to absolute safety measurements.

\paragraph{Attack Success Rate (ASR).}
The standard metric for evaluating adversarial attacks on VLA models, ASR measures the proportion of attack attempts that successfully induce the target model to produce unsafe or incorrect behavior:
\begin{equation}
\text{ASR} = \frac{|\{i : \text{attack } i \text{ succeeds}\}|}{|\mathcal{A}_{\text{attack}}|}.
\end{equation}
VLA-Risk~\cite{ru2026vlarisk} reports ASR across different modality--dimension combinations, while RoboPAIR~\cite{robey2025jailbreaking} demonstrates ASR values reaching 100\% for jailbreak attacks on LLM-controlled robots.

\paragraph{Performance Drop Rate (PDR).}
To measure the impact of adversarial perturbations relative to clean conditions, the performance drop rate is defined as:
\begin{equation}
\text{PDR} = \frac{\text{SR}_{\text{clean}} - \text{SR}_{\text{perturbed}}}{\text{SR}_{\text{clean}}} \times 100\%.
\end{equation}
VLATest~\cite{wang2025vlatest} uses this metric to quantify the sensitivity of VLA models to individual perturbation factors such as lighting, camera pose, and object distractors.

\paragraph{Certified Robustness Radius.}
Emerging from the adversarial machine learning literature, certified robustness provides a provable lower bound on the perturbation magnitude required to change a model's output.
While certified robustness methods have been applied to image classifiers and language models, their adaptation to VLA models remains an open challenge due to the sequential and multi-modal nature of robotic decision-making.

\subsubsection{Composite and Deployment-Oriented Metrics}
\label{eval_2_4}

Real-world deployment demands holistic metrics that integrate multiple safety dimensions and account for practical considerations such as economic cost and operational constraints.

\begin{wrapfigure}{r}{0.44\columnwidth}
  \centering
  \vspace{-4mm}
  \includegraphics[width=\linewidth]{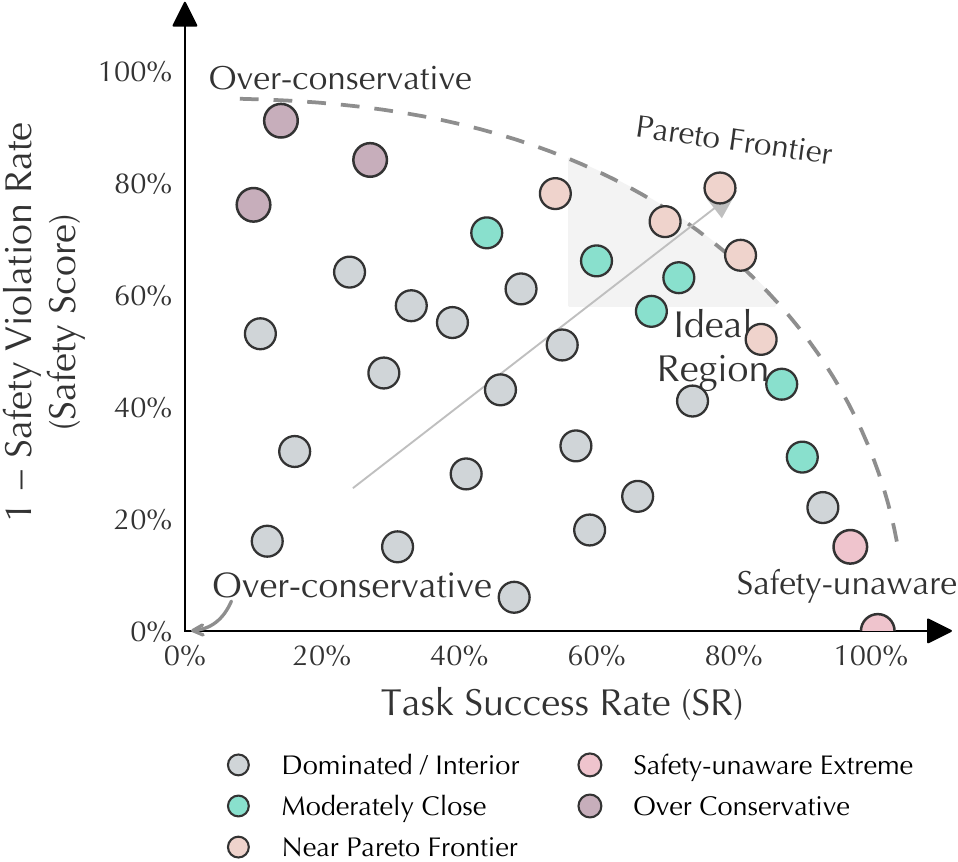}
  \vspace{-4mm}
  \caption{\textbf{Conceptual illustration of the safety--performance Pareto frontier.} Each point represents a model's operating point in the (Task Success Rate, Safety Score) space. The ideal region (upper-right) achieves both high task performance and high safety.}
  \vspace{-15mm}
  \label{fig:pareto_frontier}
\end{wrapfigure}

\paragraph{Safety--Performance Trade-off.}
A recurring theme across benchmarks is the tension between safety and task performance.
This trade-off can be visualized as a Pareto frontier in the (SR, SVR) or (SR, RejR) space, where ideal models occupy the upper-left region (high success, low violation), as conceptually illustrated in \cref{fig:pareto_frontier}.
Current evaluations consistently reveal that improving safety awareness (e.g., higher RejR) often comes at the cost of reduced task success, and finding the optimal operating point remains an open problem.

\paragraph{Cost-Aware Evaluation.}
CostNav~\cite{seong2025costnav} proposes evaluating navigation agents through a cost-revenue framework where:
\begin{equation}
\text{Net Value} = R_{\text{task}} - \sum_{j} w_j \cdot C_j,
\end{equation}
where $R_{\text{task}}$ is the revenue from task completion, $C_j$ represents different cost categories (collision damage, cargo spillage, time penalty, etc.), and $w_j$ are cost weights.
This formulation captures real-world deployment concerns that binary safety metrics miss, such as the difference between a minor scrape and catastrophic equipment damage.

\paragraph{Multi-Level Diagnostic Metrics.}
AgentSafe's Safe-Diagnose protocol~\cite{ying2025agentsafe} evaluates safety at each stage of the agent pipeline independently, measuring perception accuracy for hazard detection, planning safety for constraint satisfaction, and execution safety for physical outcome quality.
This multi-level diagnostic approach reveals that the primary vulnerability of current embodied agents lies in the \emph{planning stage}, where unsafe instructions can bypass safety filters even when the perception module correctly identifies hazards~\cite{ying2025agentsafe, chen2025safemind}.

\paragraph{Temporal Persistence of Attacks.}
For jailbreak and adversarial attacks, a critical but often overlooked metric is the \emph{temporal persistence}, i.e., how long the effects of a successful attack persist across subsequent timesteps.
One of the previous work~\cite{jones2025adversarial} demonstrates that textual jailbreak attacks on VLA models often persist over extended horizons, suggesting that single-timestep safety metrics may significantly underestimate the true risk of adversarial manipulation.

\begin{table}[t]
\centering
\caption{\textbf{Summary of safety evaluation metrics for VLA and embodied AI systems.} We organize them by task-level, behavioral, robustness, and composite perspectives. These metrics reflect the need to assess embodied safety beyond simple success rates, capturing violations, robustness degradation, and safety--performance trade-offs.}
\label{tab:safety_metrics}
\resizebox{0.65\columnwidth}{!}{%
\begin{tabular}{lll}
\toprule
\textbf{Category} & \textbf{Metric} & \textbf{Description} \\
\midrule
\multirow{3}{*}{\textbf{Task-Level}} & SVR & Proportion of episodes with $\geq$1 safety violation \\
& RejR & Rate of correctly refusing hazardous instructions \\
& SR & Task success rate (safety--performance context) \\
\midrule
\multirow{3}{*}{\textbf{Behavioral}} & CR & Frequency of unintended physical contacts \\
& SS & Multi-level perception--planning--execution score \\
& SPL & Success weighted by path efficiency \\
\midrule
\multirow{4}{*}{\textbf{Robustness}} & ASR & Proportion of successful adversarial attacks \\
& PDR & Relative performance drop under perturbation \\
& CRR & Certified lower bound on robustness \\
& LGS & Language--action sensitivity under contradictory instructions \\
\midrule
\multirow{3}{*}{\textbf{Composite}} & Net Value & Cost-revenue analysis for deployment \\
& Pareto (SR, SVR) & Safety--performance trade-off frontier \\
& Temporal Persistence & Duration of attack effects over time \\
\bottomrule
\end{tabular}%
}
\end{table}

\section{Real-World Deployment Scenarios}
\label{real_world}

The preceding sections have examined VLA safety from primarily technical perspectives, including attack vectors, defense mechanisms, and evaluation protocols.
However, the safety implications of VLA failures are inherently deployment-dependent: the same hallucinated object, unsafe instruction, or misgrounded action may lead to very different consequences in a vehicle, a home, a factory, or a hospital.
In this section, we therefore shift from method-level analysis to deployment-driven safety requirements, focusing on how physical risk, human proximity, supervision level, and regulatory constraints shape the design and evaluation of safe VLA systems.

Rather than exhaustively surveying all robotic application areas, we focus on representative domains where VLA models introduce qualitatively new safety concerns: autonomous driving (Section~\ref{real_world_1}), household robotics (Section~\ref{real_world_2}), industrial manufacturing (Section~\ref{real_world_3}), healthcare and assistive robotics (Section~\ref{real_world_4}), service and public-space robotics (Section~\ref{real_world_5}), and outdoor or field deployment (Section~\ref{real_world_6}).
We then synthesize cross-domain safety challenges in Section~\ref{real_world_7}.
Table~\ref{tab:deployment_domains} provides a comparative summary.

\subsection{Autonomous Driving}
\label{real_world_1}

Autonomous driving is one of the most safety-critical domains for VLA deployment because perception, reasoning, and action must be coupled under real-time physical constraints.
Recent surveys have observed a growing convergence between large vision-language models and end-to-end driving systems, where multimodal reasoning is used to support scene understanding, decision making, and planning~\cite{jiang2025survey, hu2025vision}.
Representative driving VLA systems, including DriveVLM~\cite{tian2024drivevlm}, OpenDriveVLA~\cite{zhou2026opendrivevla}, CoVLA~\cite{arai2025covla}, and EMMA~\cite{hwang2024emma}, demonstrate how vision-language reasoning can be integrated with trajectory prediction, planning, or action generation.
Some systems further adopt dual-system designs that separate high-level semantic reasoning from low-level reactive control, reflecting the need to balance deliberative planning with latency-sensitive execution~\cite{hu2025vision}.

The central safety issue in driving is not merely whether VLA models improve perception or planning accuracy, but whether their language-conditioned reasoning can be trusted under high-speed, high-consequence conditions.
First, VLA models inherited from vision-language foundations may hallucinate non-existent objects or misinterpret traffic semantics, potentially causing unsafe decisions such as phantom braking or failure to detect pedestrians~\cite{tian2024drivevlm, jiang2025survey}.
Second, the computational cost of large VLA models creates a latency--safety trade-off: in scenarios such as sudden pedestrian crossings or cut-in vehicles, a correct decision may still be unsafe if produced too late~\cite{hu2025vision}.
Third, driving policies must satisfy explicit traffic rules and physical constraints, motivating approaches such as SafeAuto~\cite{zhang2025safeauto}, which translates traffic rules into formal constraints for verifying predicted actions.
Safety-aware learning frameworks such as VL-SAFE~\cite{qu2025vl} and VLM-RL~\cite{huang2025vlm} further suggest that VLM-derived safety scores or semantic rewards can help guide policy optimization without relying on unsafe online trial-and-error.
Finally, RoboPAIR~\cite{robey2025jailbreaking} on NVIDIA's Dolphins self-driving LLM indicates that driving VLA systems inherit adversarial vulnerabilities from general-purpose VLA models, with the added risk that successful attacks can directly endanger human lives.

\subsection{Household and Domestic Robotics}
\label{real_world_2}

Household robotics is a representative deployment setting for VLA models because it combines open-world visual variation, ambiguous natural-language instructions, long-horizon manipulation, and close interaction with non-expert users.
Recent generalist robot policies and language-conditioned manipulation systems, including RT-2~\cite{zitkovich2023rt}, OpenVLA~\cite{kim2024openvla}, $\pi_0$~\cite{black2024pi_0}, Octo~\cite{team2024octo}, $\pi_{0.5}$~\cite{intelligence2025pi_}, SayCan~\cite{ahn2022can}, Code as Policies~\cite{liang2023code}, VoxPoser~\cite{huang2023voxposer}, and Mobile ALOHA~\cite{fu2024mobile}, indicate that language-conditioned robotic control is becoming increasingly feasible in domestic and manipulation-oriented settings.
Among these systems, the $\pi_0/\pi_{0.5}$ line is particularly relevant to household deployment because it explicitly targets messy real-world environments and demonstrates open-world generalization in kitchens and bedrooms~\cite{black2024pi_0, intelligence2025pi_}.
%Mobile ALOHA further suggests that broad household task repertoires and direct human--%robot interaction, including handover scenarios, are becoming increasingly %practical~\cite{fu2024mobile}.
Related embodied platforms such as Mobile ALOHA also suggest that broad household task repertoires and direct human--robot interaction, including handover scenarios, are becoming increasingly practical~\cite{fu2024mobile}.

The safety challenges in household deployment arise less from the nominal task categories themselves than from the unstructured and human-centered nature of homes.
Unlike controlled laboratories or factory floors, domestic environments contain changing furniture layouts, misplaced objects, pets, children, elderly users, sharp tools, hot surfaces, cleaning chemicals, and fragile items.
VLA models must therefore generalize beyond memorized training scenes, yet current models still exhibit strong tendencies toward memorization over systematic generalization~\cite{zhang2025vla, zhou2025libero}.
This creates a mismatch between benchmark success and real household safety, where a visually plausible but physically inappropriate action may cause harm.

A second challenge is safe interaction with vulnerable users.
Minor manipulation errors, such as dropping heavy objects, applying excessive force during handovers, or colliding with a person, become more serious when the interaction partner is a child, an elderly person, or a mobility-impaired user.
Although Mobile ALOHA demonstrates the feasibility of household-scale mobile manipulation and handover tasks~\cite{fu2024mobile}, ensuring safety during such close physical interaction remains an open problem.

A third challenge is hazard-aware long-horizon planning.
SafeAgentBench~\cite{yin2024safeagentbench} shows that even explicitly hazardous instructions, such as placing bread on a stove and turning it on, may not be reliably rejected by state-of-the-art agents, revealing a gap between instruction following and safety awareness.
Similarly, AgentSafe~\cite{ying2025agentsafe} shows that agents may perceive hazards but fail to translate such awareness into safe plans over extended action sequences.
Failure recovery is therefore essential: FailSafe~\cite{lin2025failsafe} demonstrates that generating diverse failure cases paired with corrective actions can improve models such as $\pi_0$-FAST and OpenVLA, suggesting that failure awareness is a learnable component of safe household autonomy.

\subsection{Industrial Manufacturing}
\label{real_world_3}

Industrial manufacturing provides a contrasting deployment setting: environments are often more structured than homes, but the physical consequences of errors are substantially higher.
The key safety question is not simply whether VLA models can support flexible assembly, inspection, or human--robot collaboration, but whether language-conditioned and visually grounded decisions can be constrained by safety-certified control layers.
Recent work on embodied AI for smart robotic cells~\cite{gupta2025embodied} suggests that foundation model-based systems may improve adaptability in dynamic factory settings, but such adaptability must be reconciled with the strict safety and reliability requirements of industrial automation.

Industrial robots operate with forces, speeds, and payloads that can cause severe injury or death.
A VLA model that misinterprets a language instruction, hallucinates an object, or generates an unexpected motion plan could direct a manipulator into a human operator's workspace.
This is especially problematic because industrial robotics is governed by safety standards such as ISO~10218 and ISO/TS~15066, which impose requirements on force limiting, speed monitoring, and safety-rated stops.
VLA models therefore cannot be deployed as unconstrained end-to-end controllers; they must interface with certified safety mechanisms, supervisory control systems, and emergency stop architectures.

Human--robot collaboration introduces an additional layer of complexity.
In collaborative manufacturing, humans and robots may share workspaces, exchange objects, or jointly manipulate tools, requiring VLA models to reason about both task progress and human safety.
Safe reinforcement learning approaches~\cite{wachi2024survey} provide one route for incorporating constraints into robot control, but adapting such constraints to multimodal VLA reasoning remains an open challenge.
The planning-level failures identified by AgentSafe~\cite{ying2025agentsafe}, where agents perceive hazards but fail to plan around them, are particularly dangerous in industrial settings where the margin for error is small.

Reliability and reproducibility are also central.
Manufacturing processes demand consistent behavior, whereas VLA outputs may be stochastic and sensitive to visual or linguistic perturbations.
The sensitivity documented by VLATest~\cite{wang2025vlatest} therefore raises concerns about whether VLA-controlled systems can satisfy the repeatability expected in industrial automation.
This tension between flexible foundation-model behavior and deterministic safety certification remains one of the core barriers to industrial VLA deployment.

\subsection{Healthcare and Assistive Robotics}
\label{real_world_4}

Healthcare and assistive robotics represent high-stakes VLA deployment settings where failures may directly affect patient well-being.
Recent systems such as RoboNurse-VLA~\cite{li2025robonurse} and Surgical-LVLM~\cite{wang2024surgical} suggest that vision-language reasoning can support surgical assistance, instrument recognition, grounded question answering, and context-aware interaction in clinical environments.
However, the safety requirements in healthcare differ sharply from general manipulation benchmarks: errors are less tolerable, users may be physically vulnerable, and regulatory requirements are substantially stricter.

In surgical and clinical settings, the tolerance for errors approaches zero.
A VLA model that misidentifies a surgical instrument, follows an ambiguous command, or executes a handover with incorrect force could cause direct patient harm.
Unlike household or service settings, where some failed actions can be retried, surgical or clinical errors may be irreversible.
This makes hallucination, misgrounding, and unsafe action generation especially concerning in healthcare VLA systems.

Healthcare environments also impose domain-specific constraints that complicate deployment.
Surgical settings require sterility, controlled interaction protocols, and reliable perception under occlusion, specialized lighting, and limited camera viewpoints.
Assistive robots, in contrast, may operate in homes or care facilities around elderly or disabled users, requiring prolonged close-contact interaction and continuous safety awareness.
The deficits in safety-aware planning revealed by SafeAgentBench~\cite{yin2024safeagentbench} and SafeMind~\cite{chen2025safemind} are particularly concerning here, because vulnerable users cannot always be expected to intervene when a robot follows an unsafe plan.

Finally, medical and assistive robots face stringent regulatory requirements.
Medical devices are subject to frameworks such as FDA 510(k) clearance and CE marking under the EU MDR, while VLA models are often stochastic, opaque, and difficult to verify.
This creates a mismatch between current certification processes, which assume traceable and testable system behavior, and learned VLA policies whose decisions may be difficult to explain or reproduce.
As a result, healthcare deployment requires not only higher task performance, but also stronger traceability, human oversight, and runtime safety assurance.

\subsection{Service and Public-Space Robotics}
\label{real_world_5}

Service and delivery robots operate in public or semi-public spaces, including sidewalks, shopping malls, hospitals, hotels, and retail environments.
Compared with household robots, they interact with a broader population of untrained users; compared with industrial robots, they face less structured and less controllable environments.
The safety challenge is therefore not only physical collision avoidance, but also socially compliant behavior, robustness to public uncertainty, and resilience against intentional or accidental interference.

Social navigation is a central requirement in this setting.
VLM-Social-Nav~\cite{song2024vlm} shows that vision-language models can provide contextual cost terms for socially appropriate robot behavior, improving success rates and reducing collision rates in social navigation scenarios.
LLM-based task planning for service robots~\cite{bian2026large} further suggests that language models can decompose high-level service requests into executable action sequences while considering environmental and social constraints.
However, these capabilities also introduce safety risks when social norms, pedestrian intent, or scene context are misinterpreted.

Public-space deployment also requires more nuanced safety evaluation than binary collision metrics.
For example, a robot may avoid collision but still move too fast in a crowded corridor, turn sharply while carrying fragile items, or block a wheelchair user's path.
CostNav~\cite{seong2025costnav} is relevant in this context because it captures cost-aware navigation behaviors that better reflect the safety implications of service robot motion.

A further concern is adversarial or opportunistic manipulation by bystanders.
Unlike household or factory robots, service robots may operate around strangers who can intentionally modify the environment, issue misleading instructions, or exploit visual-language grounding failures.
The environmental jailbreak attacks explored by Shawshank~\cite{li2025shawshank}, where hazardous behaviors are induced through environmental cues rather than explicit malicious prompts, are therefore directly relevant.
Finally, service robots frequently encounter out-of-distribution obstacles, unusual layouts, and ambiguous social situations.
The severe degradation under difficult OOD conditions documented by VLATest~\cite{wang2025vlatest} highlights the need for graceful degradation, uncertainty-aware fallback behavior, and safe refusal mechanisms in public-space VLA deployment.

\subsection{Outdoor and Field Deployment}
\label{real_world_6}

Outdoor and field deployment covers agricultural robots, outdoor delivery systems, inspection robots, and other VLA-controlled agents that operate in unstructured environments with limited human supervision.
Compared with indoor domains, these settings expose VLA models to stronger environmental variation, including direct sunlight, shadows, rain, dust, mud, terrain changes, sensor degradation, and long operating horizons.
Agricultural vision-language systems have shown potential for tasks requiring fine-grained crop understanding and domain knowledge, such as identifying growth stages or supporting treatment decisions~\cite{zhu2025harnessing}.
However, the current body of VLA-specific evidence in agriculture remains relatively limited, so this domain is best viewed as a stress case for outdoor robustness rather than a mature standalone VLA deployment area.

The main safety concern in field deployment is that perception or grounding errors can have delayed and distributed consequences.
For example, an agricultural robot that misidentifies crops, weeds, or treatment regions may incorrectly apply pesticides, herbicides, or fertilizers, causing crop damage or environmental contamination.
Such failures may not be immediately visible but can accumulate over time.
Moreover, outdoor deployment amplifies robustness challenges already observed in VLA benchmarks: the sensitivity to visual perturbations documented by VLATest~\cite{wang2025vlatest} and VLA-Risk~\cite{ru2026vlarisk} suggests that current models may be insufficiently reliable under sustained lighting, weather, and terrain variation.
Because field robots may operate for extended periods with minimal supervision, the temporal persistence of adversarial effects documented by Jones et al.~\cite{jones2025adversarial} is especially concerning.

\subsection{Cross-Domain Safety Challenges}
\label{real_world_7}
Across deployment domains, several safety challenges recur despite differences in physical setting, supervision level, and regulatory maturity.

\paragraph{The simulation-to-reality gap.}
The majority of existing safety benchmarks (Table~\ref{tab:safety_benchmarks}) operate in simulation environments.
While simulation enables systematic and repeatable evaluation, safety guarantees established in simulation do not automatically transfer to physical deployment.
Sensor noise, mechanical wear, communication latency, actuator uncertainty, and environmental variability can introduce failure modes that are absent or inadequately modeled in benchmarks~\cite{lu2025exploring}.
This gap is particularly important for VLA models because perception, language grounding, and action execution are coupled: a small visual or physical deviation can propagate into incorrect semantic interpretation and unsafe action generation.

\begin{figure*}[t]
  \centering
  \includegraphics[width=0.75\textwidth]{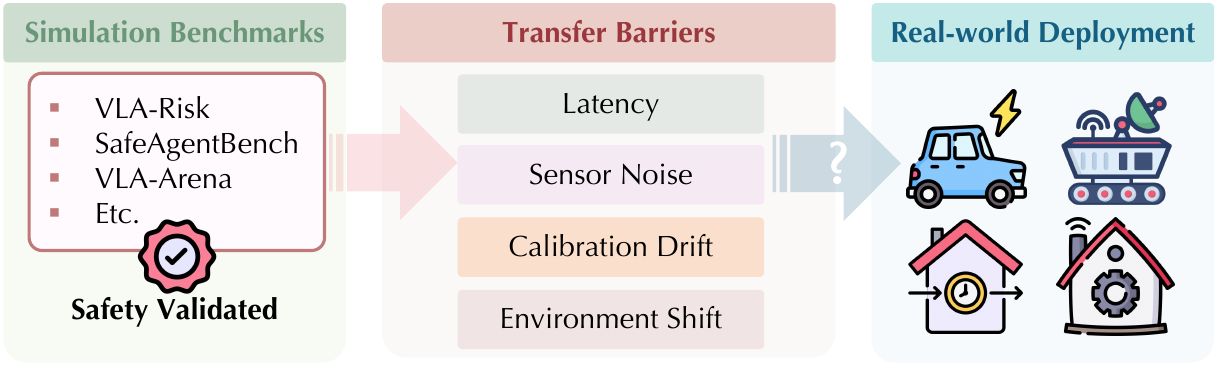}
  \caption{\textbf{The simulation-to-reality gap in VLA safety evaluation.} Safety guarantees established through simulation benchmarks (left) must transfer across multiple gap factors before they can be relied upon in real-world deployment (right). Current benchmarks predominantly operate in simulation, leaving the transferability of safety assurances as a fundamental open challenge.}
  \label{fig:sim2real_gap}
  \vspace{-2mm}
\end{figure*}

\paragraph{Scalable runtime verification and monitoring.}
As VLA models are deployed across more diverse physical scenarios, exhaustive pre-deployment testing becomes infeasible.
The space of possible instructions, visual observations, human behaviors, and environmental states is combinatorially large.
Therefore, safe deployment requires scalable runtime monitors, formal or semi-formal constraints, and fallback mechanisms that can continuously check whether VLA-generated plans remain within acceptable safety boundaries.

\paragraph{The safety--capability Pareto frontier.}
A recurring theme across domains is the tension between task performance and safety.
Overly conservative robots may refuse too often or become practically unusable, while aggressive policies may achieve higher task success at the cost of unsafe behavior.
The appropriate operating point is domain-dependent: a household robot may prioritize caution around vulnerable users, an autonomous vehicle must balance rapid response with rule compliance, and a surgical assistant may require near-zero tolerance for unsafe actions.

\paragraph{Regulation, accountability, and safety drift.}
Traditional safety certification processes often assume deterministic, verifiable system behavior, whereas VLA models are typically stochastic, opaque, and sensitive to distribution shifts.
Although emerging regulatory frameworks such as the EU AI Act have begun to address high-risk AI systems, specific guidance for VLA-controlled physical systems remains underdeveloped.
This problem becomes more difficult when VLA models are updated after deployment through fine-tuning, human feedback, online learning, or data augmentation.
Such updates may introduce \emph{safety drift}, where previously safe behaviors are inadvertently changed.
Maintaining safety over the deployment lifecycle therefore requires continuous monitoring, regression testing, and clear accountability mechanisms.

\paragraph{Fleet-level and multi-agent safety.}
As VLA-controlled robots are deployed in fleets, such as warehouse robots, delivery vehicles, or coordinated service robots, safety concerns extend beyond individual agents.
A failure in one agent may affect others through shared plans, communication assumptions, or physical interactions.
Future safety frameworks therefore need to consider not only single-agent instruction following, but also coordination, shared situational awareness, and cascading failures across multiple embodied agents.

\begin{table*}[t]
\centering
\caption{\textbf{Comparative summary of VLA deployment domains and their safety characteristics.} ``Severity'' indicates the potential consequence of safety failures; ``Supervision'' indicates the typical level of human oversight during operation; ``Regulatory'' indicates the maturity of applicable safety regulations.}
\label{tab:deployment_domains}
\resizebox{\textwidth}{!}{%
\begin{tabular}{lllllll}
\toprule
\textbf{Domain} & \textbf{Representative Systems} & \textbf{Key Safety Risks} & \textbf{Severity} & \textbf{Supervision} & \textbf{Regulatory} & \textbf{Environment} \\
\midrule
\textbf{Autonomous Driving} & DriveVLM, EMMA, SafeAuto & Collision, pedestrian harm, delayed response & Critical & Low--Medium & Mature & Structured outdoor \\
\textbf{Household Robotics} & RT-2, OpenVLA, $\pi_0/\pi_{0.5}$ & Object mishandling, hazardous actions, human contact & Moderate--High & Low & Emerging & Unstructured indoor \\
\textbf{Industrial Manufacturing} & Smart robotic cells, collaborative VLA systems & Worker injury, equipment damage, unsafe collaboration & Critical & Medium--High & Mature & Structured indoor \\
\textbf{Healthcare \& Assistive} & RoboNurse-VLA, Surgical-LVLM & Patient harm, surgical errors, unsafe assistance & Critical & High & Strict & Controlled indoor \\
\textbf{Service \& Public-Space} & VLM-Social-Nav, CostNav & Pedestrian collision, social non-compliance, public interference & Moderate & Low & Emerging & Semi-structured public \\
\textbf{Outdoor \& Field} & Agricultural VLA systems & Environmental contamination, robustness failure, low supervision & Moderate & Very Low & Developing & Unstructured outdoor \\
\bottomrule
\end{tabular}%
}
\vspace{-4mm}
\end{table*}
\section{Future Directions}
\label{future}

The preceding sections have mapped the threat surface, defense repertoire, evaluation protocols, and deployment landscape of VLA safety.
Despite substantial progress, the field remains at an early stage: most known attacks have no principled countermeasures, most defenses lack formal guarantees, and most evaluations are confined to simulation.
In this section, we outline research directions that we view as both urgent and promising, organized around five themes that recur across attack, defense, evaluation, and deployment literatures.

\subsection{Certified Robustness and Physically Realizable Defenses}

Existing VLA defenses are predominantly empirical, validated against a fixed catalogue of attacks without formal guarantees over the full perturbation space.
Translating certified robustness tools from image classification to VLA raises unresolved difficulties: the perturbation space is cross-modal, with visual, linguistic, proprioceptive, and physical channels that can be attacked jointly; certificates must cover entire trajectories rather than individual frames, since small per-step errors compound; and they must respect real-time constraints that rule out expensive methods such as full Lipschitz-bounded architectures at test time.
Progress will likely require new notions of certification that combine per-step bounds with trajectory-level stability.

A parallel concern is the gap between studied and deployed threat models.
Much of the attack literature studies digital perturbations---pixel-space noise, in-image patches, or prompt manipulations---while real deployments face physically realizable interventions: printed patches, object substitutions, lighting manipulation, and acoustic injection~\cite{li2025shawshank, robey2025jailbreaking}.
Defensive research must therefore incorporate physically rendered adversarial data at training time and combine multi-view consensus, proprioceptive cross-checks, and language-grounded plausibility verification at inference time.
The multi-modal attack surface, often viewed as a liability, becomes a defensive asset: the more modalities an attacker must manipulate consistently, the harder an undetected attack becomes.

\subsection{Safety-Aware Training and Unified Runtime Architectures}

Current VLA training pipelines are dominated by behavior cloning on demonstrations where safety is implicit rather than explicit, leaving the learner vulnerable to the data poisoning and backdoor attacks surveyed in Section~\ref{train_attack}.
Several paradigms deserve deeper investigation:
safety-constrained policy optimization that encodes safety as explicit constraints during fine-tuning~\cite{wachi2024survey};
constitutional and red-team-driven alignment that transfers language-model alignment practices to embodied settings;
curriculum-based safety training that pedagogically sequences safe and unsafe scenarios;
and human-in-the-loop refinement that distills expert corrections into the policy via preference learning.
Integrating these paradigms without sacrificing generalization---VLA systems derive much of their value from broad behavioral coverage---remains a central methodological challenge.

At inference time, decision-layer guardrails, closed-loop monitors, and physical fail-safes (Section~\ref{infer_defense}) are presently studied in isolation.
A unified runtime architecture must allocate computation adaptively to meet heterogeneous latency budgets (sub-100~ms for autonomous driving, longer for household manipulation), arbitrate conflicting interventions when layered defenses issue contradictory commands, and produce useful safety signals even under interruption (\emph{anytime safety}).
Promising building blocks combine control-theoretic tools---control barrier functions, reachability analysis, signal temporal logic---with learned uncertainty estimates from the VLA itself, as in SAFE~\cite{gu2025safe} and SAFE-SMART~\cite{sakano2025safe}.

\subsection{Standardized Evaluation and Sim-to-Real Transfer}

Section~\ref{eval} surveyed a fragmented benchmark landscape with heterogeneous scenarios, metrics, and assumptions, making it difficult to compare defenses, track progress, or detect regressions.
Three priorities stand out.
First, the community needs a shared \emph{safety evaluation kit} that packages representative scenarios (hazardous instructions, visual perturbations, environmental jailbreaks, long-horizon plans) with standard metrics (success rate, collision rate, rejection rate, STL satisfaction, ECE) and reference implementations.
Second, benchmarks should include explicit \emph{adversarial} splits that stress-test models under worst-case inputs rather than nominal conditions.
Third, benchmarks should target the \emph{safety--capability frontier}, reporting both safety and task performance so that overly conservative models cannot hide behind high rejection rates.

A related challenge, highlighted in Section~\ref{real_world_7}, is that most current safety evaluations operate in simulation while deployment consequences are borne in the physical world.
Bridging this gap calls for domain randomization that covers safety-critical distribution shifts (sensor noise, actuator dynamics, rare physical events), hybrid pipelines that combine large-scale simulation with targeted physical validation, and formal links---via probabilistic guarantees, distributional robustness, or conformal prediction---between simulated and real-world safety metrics.
Without such links, simulation-established guarantees offer limited assurance at deployment time.

\subsection{Lifecycle Safety: Continuous Learning and Fleet-Level Deployment}

VLA models deployed in practice are rarely static: they are fine-tuned on new data, updated to incorporate user feedback, and retrained as new tasks or embodiments are added.
Each update risks \emph{safety drift}, where previously safe behaviors degrade in ways that may be invisible until a failure occurs~\cite{wang2025vlatest, ying2025agentsafe}.
The high-dimensional input space and stochastic outputs of VLA systems make exhaustive regression testing impractical, motivating research on \emph{safety regression suites} that efficiently characterize behavioral changes, \emph{drift-aware fine-tuning} methods that explicitly preserve safety-critical behaviors, and \emph{post-deployment monitoring} that detects emerging unsafe patterns from telemetry.

Large-scale deployments---warehouse fleets, delivery networks, autonomous vehicle fleets---amplify these concerns and introduce genuinely multi-agent failure modes.
A localized failure in one unit can cascade through shared planning assumptions, physical interactions, or correlated errors induced by a common model checkpoint.
Priorities include fleet-wide monitoring that aggregates telemetry to detect systemic failures, coordination-aware safety policies that internalize how an agent's behavior affects nearby agents, and supply-chain safety for model checkpoints and training data distributed across teams.
When multiple VLA-controlled robots interact with humans in shared spaces, safety also depends on legible, predictable behavior---properties that current end-to-end models do not explicitly optimize for.

\subsection{Regulatory, Ethical, and Societal Considerations}

The technical questions above are inseparable from governance questions.
Regulatory regimes---FDA medical device clearance, ISO industrial safety standards, the EU AI Act---presume verifiable, auditable system behavior, yet VLA models are neither transparent nor deterministic.
Research is needed on \emph{auditable VLA architectures} that expose decision traces suitable for regulatory review, \emph{risk-tiered evaluation} that aligns benchmark results with regulatory categories, and \emph{liability frameworks} that allocate responsibility across model developers, integrators, and operators when failures occur.
Ethical concerns---privacy implications of always-on perception, equity in access to assistive robotics, accountability for autonomous harm---interact with these technical and regulatory threads and deserve explicit treatment rather than deferral to downstream deployments.

Viewed as a whole, these directions reflect a broader goal: turning VLA safety from a collection of ad hoc attack--defense skirmishes into a cumulative scientific discipline with shared threat models, reproducible benchmarks, and theoretical frameworks.
The embodied consequences of VLA failures leave little room for a purely post-hoc approach.
Building safety into VLA models from the outset is both a scientific challenge and a societal obligation.

\section{Conclusion}
\label{conclusion}

Vision-Language-Action models have rapidly advanced from research prototypes to systems deployed in autonomous vehicles, household assistants, industrial manipulators, surgical robots, and field machinery.
Their unified treatment of perception, language, and action enables generalization that was unattainable with earlier modular stacks, but it also reshapes the safety problem: perturbations in any modality can propagate into physical actions with consequences that cannot be undone by a software patch.
This survey has offered, to our knowledge, the first comprehensive synthesis of the emerging VLA safety literature, structured along two parallel timing axes---attack timing (training-time vs.\ inference-time) and defense timing (training-time vs.\ inference-time)---that pair each class of threat with the stage at which it can be mitigated.

Our review documented the growing catalogue of threats that VLA systems face.
On the training side (Section~\ref{train_attack}), we surveyed data poisoning, input- and state-space backdoors, and temporal action-chunking exploits that embed unsafe behaviors directly into learned policies.
On the inference side (Section~\ref{infer_time}), we examined semantic jailbreaks that bypass language-level safeguards, visual and cross-modal perturbations that exploit multi-modal integration, and physical interventions that compromise deployed systems through the environment itself.
We then analyzed the corresponding defensive repertoire: pedagogical data design, constrained policy optimization, and human-in-the-loop refinement on the training side (Section~\ref{train_defense}); decision-layer guardrails, closed-loop monitors, and physical fail-safes on the inference side (Section~\ref{infer_defense}).
Throughout, we emphasized constraints unique to embodied systems---the safety--latency trade-off, trajectory-level error propagation, and the high cost of conservatism---that differentiate VLA safety from its text-only counterpart.

\begin{wrapfigure}{r}{0.44\columnwidth}
  \centering
  % \vspace{-3mm}
  \includegraphics[width=\linewidth]{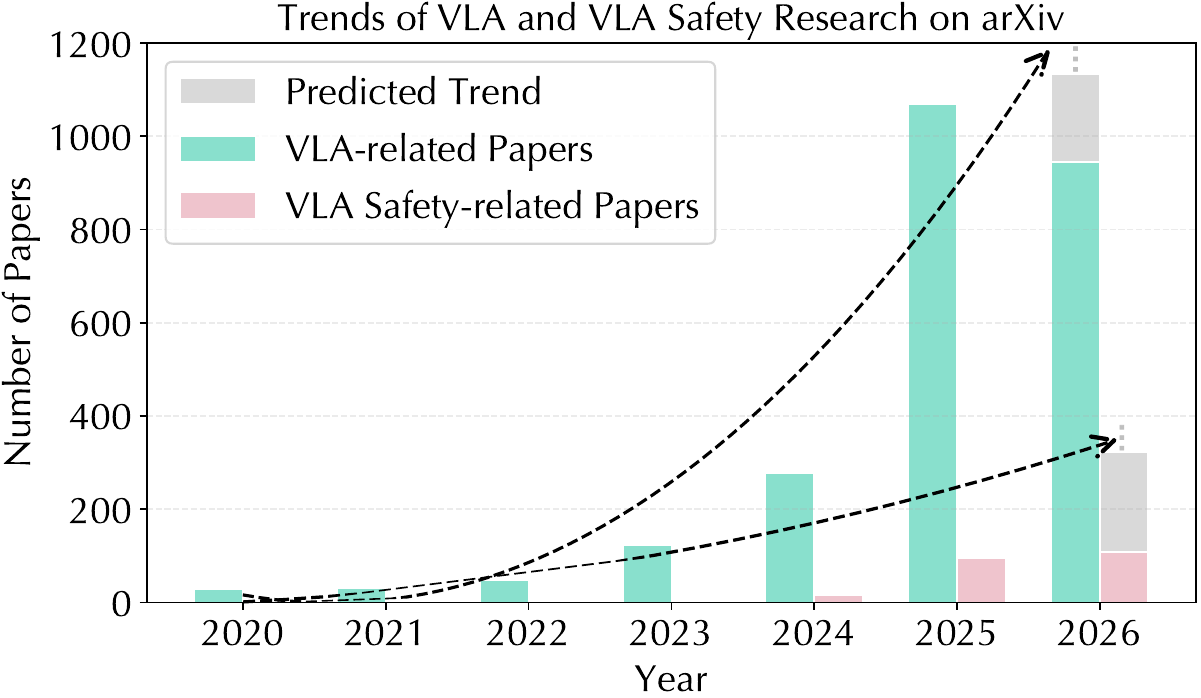}
  \vspace{-4mm}
  \caption{\textbf{Number of arXiv publications retrieved via keyword-based search (titles and abstracts) for VLA-related papers and VLA safety-related papers.} The results show a rapid increase in both general VLA research and safety-focused VLA studies, with safety-related work growing particularly sharply after 2024.}
  \vspace{-4mm}
  \label{fig:vla_safety_trend}
\end{wrapfigure}

The evaluation landscape we surveyed in Section~\ref{eval} reveals a field that is maturing unevenly, with rapid growth in both general VLA research and safety-focused studies in recent years (\cref{fig:vla_safety_trend}).
Benchmarks such as VLA-Risk, VLATest, SafeAgentBench, and AgentSafe have established first-generation coverage of training- and inference-time threats, and runtime-alignment suites such as ASIMOV, SAFE-SMART, and SAFE extend evaluation into the operational regime, but metrics remain heterogeneous, adversarial splits are uneven, and simulation dominates at the expense of The deployment analysis in Section~\ref{real_world} showed that six domains---autonomous
physical validation.
driving, household robotics, industrial manufacturing, healthcare and assistive robotics, service and delivery, and agricultural and field robotics---share many core challenges yet impose domain-specific constraints on acceptable failure modes, supervision levels, and regulatory expectations.
The cross-domain synthesis highlighted recurring themes that no single community has fully resolved: the sim-to-real gap in safety guarantees, the combinatorial difficulty of verification at scale, and the fundamental tension between safety and capability.

In Section~\ref{future}, we outlined directions that we view as most urgent: certified robustness tailored to embodied trajectories, physically realizable attack and defense research, safety-aware training paradigms, unified runtime safety architectures, standardized and reproducible evaluation, sim-to-real safety transfer, continuous learning under safety drift, fleet-level safety, and regulatory alignment.
These directions are interconnected---progress on one rarely suffices without progress on several others---and they call for collaboration across robotic learning, adversarial machine learning, control theory, and AI alignment communities that have historically operated with limited cross-pollination.

This survey has limitations worth acknowledging.
The VLA safety literature is growing rapidly, and new attacks, defenses, and benchmarks will inevitably appear after the present snapshot.
Our taxonomies, while designed to be forward compatible, may require extension as physical attacks and fleet-level threats mature.
We have emphasized breadth over depth in places where the underlying research is still nascent, and we expect individual threads---certified robustness, runtime monitoring, and regulatory integration, in particular---to warrant dedicated surveys of their own in the near future.

Notwithstanding these limitations, we believe the broad picture is clear.
VLA models are leaving the laboratory, and their safety is no longer a speculative concern.
The window in which safety can be retrofitted into deployed systems is narrow; the more responsible path is to make safety a first-class design objective alongside capability, efficiency, and generalization.
We hope this survey helps orient researchers, practitioners, and policymakers entering the field, and we will maintain the accompanying repository as a living resource to track the community's progress.
The promise of VLA systems---general-purpose embodied assistants that understand and act on human intent---will only be realized if that promise can be kept safely.

% \newpage

\bibliographystyle{assets/plainnat}
\bibliography{citation}

\end{document}